\documentclass[journal,twoside,web]{ieeecolor}
\usepackage{generic}
\usepackage{textcomp}

\usepackage{amsmath,amssymb,mathtools}
\usepackage{bm}
\usepackage{booktabs}
\usepackage{algorithm}
\usepackage{algpseudocode}
\usepackage{cite}
\usepackage[hidelinks]{hyperref}
\usepackage{microtype}
\usepackage{graphicx}
\usepackage[caption=false,font=footnotesize]{subfig}
\usepackage{array}
\usepackage{dsfont}
\usepackage{balance}
\usepackage{comment}
\usepackage{newtxtext,newtxmath}
\usepackage{tikz}
\usetikzlibrary{arrows.meta,positioning,shapes.geometric,fit,calc,backgrounds}

\newcolumntype{C}{>{\centering\arraybackslash}p{1.05cm}}

\newtheorem{lemma}{Lemma}
\newtheorem{proposition}{Proposition}

\newcommand{\vect}{\operatorname{vec}}
\newcommand{\ind}{\mathrm{I}}
\newcommand{\R}{\mathbb{R}}
\newcommand{\Ical}{\mathcal{I}}
\newcommand{\Ecal}{\mathcal{E}}
\newcommand{\Ncal}{\mathcal{N}}
\newcommand{\norm}[1]{\left\lVert #1\right\rVert}
\newcommand{\trans}{^{\mathsf T}}

\newcommand{\col}{\operatorname{col}}

\definecolor{layerblue}{RGB}{240,245,250}
\definecolor{layergray}{RGB}{248,248,248}
\definecolor{accent}{RGB}{48,82,116}
\definecolor{darkgray}{RGB}{55,55,55}
\definecolor{commred}{RGB}{150,55,55}

\def\BibTeX{{\rm B\kern-.05em{\sc i\kern-.025em b}\kern-.08em
    T\kern-.1667em\lower.7ex\hbox{E}\kern-.125emX}}
\begin{document}
\title{Graph-Sparse Distributed Predictive Control for Multi-Robot Motion Planning via Edge-Consensus ADMM}
\author{Josep Borr\`as-Borr\`as, Pablo Segovia and Carlos Ocampo-Martinez,~\IEEEmembership{Senior Member, IEEE}
\thanks{This work has been supported by the Spanish project SEAMLESS (ref. PID2023-148840OB-I00), funded by MCIN/AEI/10.13039/501100011033/FEDER, UE.}
\thanks{All authors are with the Automatic Control Department (ESAII), Universitat Polit\`ecnica de Catalunya -- BarcelonaTech, Barcelona, Spain.
{\tt\small josep.borras.borras@estudiantat.upc.edu, \{pablo.segovia, carlos.ocampo\}@upc.edu}}
\thanks{OpenAI Codex was used to assist in drafting some statements in Sections~V-C--V-E, and to support language editing and consistency checks throughout the manuscript. The authors take full responsibility for the mathematical correctness and all other content of this article.}}

\maketitle

\begin{abstract}
This paper presents a distributed model predictive control (MPC) architecture for collision-avoiding motion planning of planar robots with double-integrator dynamics. To reduce coordination demands, a time-varying predicted-conflict graph selects relevant pairwise interactions. The associated collision-avoidance constraints are convexified once per MPC update using warm-start trajectories, yielding a fixed convex quadratic program (QP). The resulting QP is then decomposed using an edge-consensus formulation of the alternating direction method of multipliers (ADMM), enabling parallel robot-local optimization and independent closed-form projections onto separating half-spaces. Execution is selectively supervised by a centralized high-order control barrier function (HOCBF) QP when the distributed candidate does not satisfy the direct-execution conditions. Evaluated in C++ simulations involving 8, 16, and 32 robots, the proposed architecture achieves collision-free execution and substantially improves computational scalability while maintaining task-level performance comparable to a centralized complete-graph baseline.
\end{abstract}


\section{Introduction}
\label{sec:introduction}

\IEEEPARstart{T}{he} growing adoption of autonomous robotic systems in logistics, autonomous driving, marine navigation, and aerial robotics has intensified the need for scalable multi-robot motion planning~\cite{Li2021Warehouse,Hult2019,Luis2020}. Coordinating multiple robots in shared environments requires planning collision-free, dynamically feasible trajectories while managing the computational and communication demands of inter-robot interactions~\cite{Yang2025}.

Model predictive control (MPC) is well suited to multi-robot motion planning because robot dynamics, actuation limits, trajectory objectives, and collision-avoidance requirements can be incorporated within a unified receding-horizon optimization problem~\cite{Yu2021}. However, centralized formulations become increasingly expensive as the fleet size and number of pairwise interactions grow. Distributed MPC alleviates this limitation by decomposing the fleet-level problem into smaller coupled subproblems~\cite{Firoozi2024}. The alternating direction method of multipliers (ADMM) is particularly suited to this structure because it can separate robot-local dynamics and objectives from inter-robot coupling constraints~\cite{boyd2011}.

Existing ADMM-based distributed MPC approaches adopt different strategies to handle nonconvex inter-robot collision-avoidance constraints. Ferranti et al.~\cite{ferranti2023} retain the nonlinear constraints within a nonlinear-ADMM framework, requiring constrained nonconvex optimization at each coordination iteration. 
In contrast, Rey et al.~\cite{rey2018} linearize the collision-avoidance constraints and update the linearization during ADMM to reduce conservatism. However, standard convergence guarantees require eventually fixing the linearization, and coordination involves constrained optimization over each robot's neighborhood.
Nguyen et al.~\cite{Nguyen2023} also employ linearized collision-avoidance constraints, constructed from previous MPC predictions, together with local copies of neighboring trajectories. However, their numerical evaluation is limited to four-robot scenarios, and their formulation retains collision-avoidance constraints for all communicating neighbors, regardless of predicted trajectory conflicts.

Beyond simplifying collision-avoidance constraints, limiting inter-robot interactions offers another means of reducing computational and communication demands.
Tajbakhsh et al.~\cite{Tajbakhsh2024} identify conflicts between predicted trajectories and incrementally introduce collision-avoidance constraints through a conflict-based search tree. Although this avoids unnecessary constraints, resolving successive conflicts may require multiple MPC reoptimizations and tree expansions. 
Serra-Gómez et al.~\cite{SerraGomez2023} employ a learned communication policy that selectively exchanges predicted trajectories based on anticipated collision risk. However, their approach reduces communication rather than the number of collision-avoidance constraints.
De Vos et al.~\cite{deVos2026} selectively activate inter-agent coupling constraints using a structured world model and overlapping planned area horizons, reducing coordination demands but relying on accurate environment-specific topological information.

Control barrier functions (CBFs) have also been employed to enforce collision-avoidance requirements in multi-robot coordination. 
Jiang and Guo~\cite{Jiang2024} incorporate discrete-time CBF constraints directly into ADMM-based distributed MPC. However, this couples collision-avoidance enforcement to the distributed optimization process, where finite ADMM iterations may leave non-negligible feasibility residuals. 
Alternatively, Tistaert et al.~\cite{tistaert2026} combine decentralized ADMM with a centralized high-order CBF (HOCBF) safety filter, separating execution-level safety supervision from the distributed optimizer. Their formulation retains the original nonconvex collision-avoidance constraints, requiring nonlinear coordination subproblems, and adopts a complete interaction graph. Furthermore, the centralized safety-filtering problem is solved at every MPC update, even when no correction to the distributed input is required.

These approaches highlight important trade-offs between collision-avoidance conservatism, computational complexity, scalability, and safety. However, the reviewed methods do not jointly combine prediction-based interaction sparsification, a fixed convexified optimization problem with inexpensive pairwise ADMM updates, finite-iteration feasibility assessment, and selective execution-level safety supervision. Motivated by these limitations, this paper develops a graph-sparse distributed MPC architecture with the following main contributions:

\begin{enumerate}
    \item A graph-sparse distributed MPC architecture that combines predicted-conflict-based interaction selection, a single warm-start-based collision-avoidance convexification per MPC update, and edge-consensus ADMM with parallel robot-local QPs and closed-form pairwise projections.

    \item A finite-iteration coordination and execution-supervision strategy that combines a geometric separation certificate under incomplete ADMM consensus with all-pairs candidate validation, first-interval clearance checking, and selective HOCBF filtering.

    \item A comparative C++ simulation study involving 8, 16, and 32 robots over 20 randomized realizations per fleet size, evaluating computational scalability, interaction sparsity, collision clearance, and task-level performance against a centralized complete-graph MPC baseline.
\end{enumerate}

The remainder of this paper is organized as follows: Section~II formulates the multi-robot MPC problem. Sections~III--V present the predicted-conflict graph, constraint convexification, and edge-consensus ADMM formulation, respectively. Section~VI describes the execution supervision layer, Section~VII presents the simulation results, and Section~VIII concludes the paper and outlines future work.

\section{Problem Statement}

\subsection{Robot model}

Consider $M$ planar robots indexed by $\Ical=\{1,\ldots,M\}$. At discrete time $k$, robot $i\in \mathcal{I}$ is described by the state and control input
\begin{equation}
    x_{i,k}=\col(p_{i,k},v_{i,k})\in\R^4,
    \qquad a_{i,k}\in\R^2,
\end{equation}
where $\col(\cdot)$ denotes vertical concatenation, and $p_{i,k}$, $v_{i,k}$, and $a_{i,k}$ denote its planar position, velocity, and acceleration, respectively. 

Assuming zero-order-hold inputs over a sampling interval $T_s$, the exact discrete-time double-integrator dynamics are
\begin{subequations}\label{eq:disc_dyn}
\begin{align}
    p_{i,k+1} &= p_{i,k}+T_s v_{i,k}+\tfrac{1}{2}T_s^2 a_{i,k},\\
    v_{i,k+1} &= v_{i,k}+T_s a_{i,k}.
\end{align}
\end{subequations}

Componentwise velocity and planning-layer acceleration bounds are imposed as \begin{equation}\label{eq:box_limits} \norm{v_{i,k}}_\infty\leq v_i^{\max}, \qquad \norm{a_{i,k}}_\infty\leq a_i^{\max}. \end{equation}
These box constraints preserve the linear structure of the local MPC constraints and hence the QP structure used later in the distributed formulation.

Each robot is modeled as a disk of radius $r_i$. For a robot pair $(i,j)$, the prescribed minimum center-to-center distance is \begin{equation}\label{eq:dij} d_{ij}^{\min} =r_i+r_j+d_{\mathrm{marg}}, \end{equation} where $d_{\mathrm{marg}}\geq0$ denotes an additional safety margin.

\subsection{Path-progress parametrization}

For each robot $i$, the reference path is the straight segment connecting its initial position $p_i^{\mathrm s}$ to its goal $p_i^{\mathrm g}$. Assuming $p_i^{\mathrm s}\neq p_i^{\mathrm g}$, define path length, unit tangent, and unit normal as 
\begin{equation}
    L_i=\norm{p_i^{\mathrm g}-p_i^{\mathrm s}}_2,
    \,
    t_i=\frac{p_i^{\mathrm g}-p_i^{\mathrm s}}{L_i},
    \,
    n_i=\begin{bmatrix}-t_{i,y}&t_{i,x}\end{bmatrix}\trans.
\end{equation}

Within each MPC problem, let $\ell\in\{0,\ldots,N\}$ denote the
prediction stage, with dependence on the current MPC sample omitted for
compactness. Following the virtual-progress parametrization used in model predictive contouring control (MPCC)~\cite{liniger2015}, a progress coordinate $\phi_{i,\ell}\in[0,L_i]$ parametrizes the reference point along the path as
\begin{equation}
p_{i,\ell}^{\mathrm{ref}}
=
p_i^{\mathrm s}+t_i\phi_{i,\ell}.
\end{equation}

For $\ell=0,\ldots,N-1$, its evolution is governed by
\begin{equation}
\label{eq:phi_dyn}
\phi_{i,\ell+1}
=
\phi_{i,\ell}+T_s\nu_{i,\ell},
\qquad
0\leq\nu_{i,\ell}\leq\nu_i^{\max}.
\end{equation}

To make the virtual progress rate consistent with the componentwise velocity limits in~\eqref{eq:box_limits}, its upper bound is chosen as 
\begin{equation} \nu_i^{\max} = v_i^{\max}\norm{t_i}_1 = v_i^{\max}\left(|t_{i,x}|+|t_{i,y}|\right), 
\end{equation} 
which is the maximum tangential velocity attainable under
$\norm{v_i}_\infty\leq v_i^{\max}$.

The augmented state and input used in the local MPC problem are therefore
\begin{equation}
\begin{aligned}
    \xi_{i,\ell} &= \col(p_{i,\ell},v_{i,\ell},\phi_{i,\ell})\in\R^5,\\
    u_{i,\ell} &= \col(a_{i,\ell},\nu_{i,\ell})\in\R^3.
\end{aligned}
\end{equation}
The corresponding predicted state and input sequences are denoted by
\begin{equation}
\label{eq:state_input}
    \xi_i=\{\xi_{i,\ell}\}_{\ell=0}^{N},
    \qquad
    u_i=\{u_{i,\ell}\}_{\ell=0}^{N-1}.
\end{equation}

\subsection{Local MPC objective}
\label{sec:localMPCobj}

The path-following objective is defined through the contouring and longitudinal errors \begin{subequations}\label{eq:path_errors} \begin{align} e_{i,\ell}^{\mathrm c} &=n_i\trans(p_{i,\ell}-p_i^{\mathrm s}),\\ e_{i,\ell}^{\mathrm l} &=\phi_{i,\ell}-t_i\trans(p_{i,\ell}-p_i^{\mathrm s}). 
\end{align} 
\end{subequations}

The contouring error measures the lateral deviation from the reference path, whereas the longitudinal error couples the virtual progress coordinate to the physical position of the robot.

For a prediction horizon $N\ge2$, the local objective is
\begin{multline}
    J_i={}\sum_{\ell=1}^{N-1}
    \left(q_{\mathrm c}(e_{i,\ell}^{\mathrm c})^2
    +q_{\mathrm l}(e_{i,\ell}^{\mathrm l})^2\right)
    +\sum_{\ell=0}^{N-1}a_{i,\ell}\trans R_a a_{i,\ell}\\
    +\sum_{\ell=0}^{N-1}w_{\ell}
    \norm{a_{i,\ell}-\bar a_{i,\ell}}_2^2
    +J_{i,N},
    \label{eq:local_cost}
\end{multline}
where $\bar a_{i,\ell}$ denotes the reference acceleration obtained from the warm-start procedure described in Section~\ref{subsec:warm_start}, and
\begin{multline}
    J_{i,N}={}q_{\mathrm c}(e_{i,N}^{\mathrm c})^2
    +q_{\mathrm l}(e_{i,N}^{\mathrm l})^2\\
    +\norm{\col(p_{i,N}-p_i^{\mathrm g},v_{i,N})}_{P_f}^2
    +q_\phi(\phi_{i,N}-L_i)^2.
    \label{eq:terminal_cost}
\end{multline}

The terminal progress term $q_\phi(\phi_{i,N}-L_i)^2$ promotes advancement toward the goal, while the longitudinal-error penalty discourages the virtual progress coordinate from advancing independently of the physical position of the robot.

The weighting parameters satisfy
$q_{\mathrm c},q_{\mathrm l},q_\phi\geq0$,
$R_a\succeq0$, and $P_f\succeq0$.
The proximal weight decays along the horizon according to
\begin{equation}
\label{eq:prox_weight}
    w_\ell=w_0\beta^\ell,\qquad w_0>0,\qquad 0<\beta<1,
\end{equation}
so deviations from the warm-start acceleration sequence are penalized
more strongly for inputs closer to execution.

Let $\mathcal F_i$ denote the set of state and input sequences
satisfying the prescribed initial conditions, the dynamics and
bounds in~\eqref{eq:disc_dyn}, \eqref{eq:box_limits},
and~\eqref{eq:phi_dyn}, together with the prescribed
virtual-progress bounds.
The underlying multi-robot MPC problem is
\begin{align}
    \min_{\{\xi_i,u_i\}}
    \quad & \sum_{i\in\Ical} J_i(\xi_i,u_i)
    \label{eq:multirobot_mpc}\\
    \text{s.t.}\quad
    &(\xi_i,u_i)\in\mathcal F_i,
    \qquad i\in\Ical, \nonumber\\
    &\norm{p_{i,\ell}-p_{j,\ell}}_2
    \geq d_{ij}^{\min},
    \quad i<j,\ \ell=1,\ldots,N. \nonumber
\end{align}

\begin{figure*}[t]
    \centering
    \makebox[\textwidth][c]{%
        \resizebox{1.01\textwidth}{!}{%
            \input{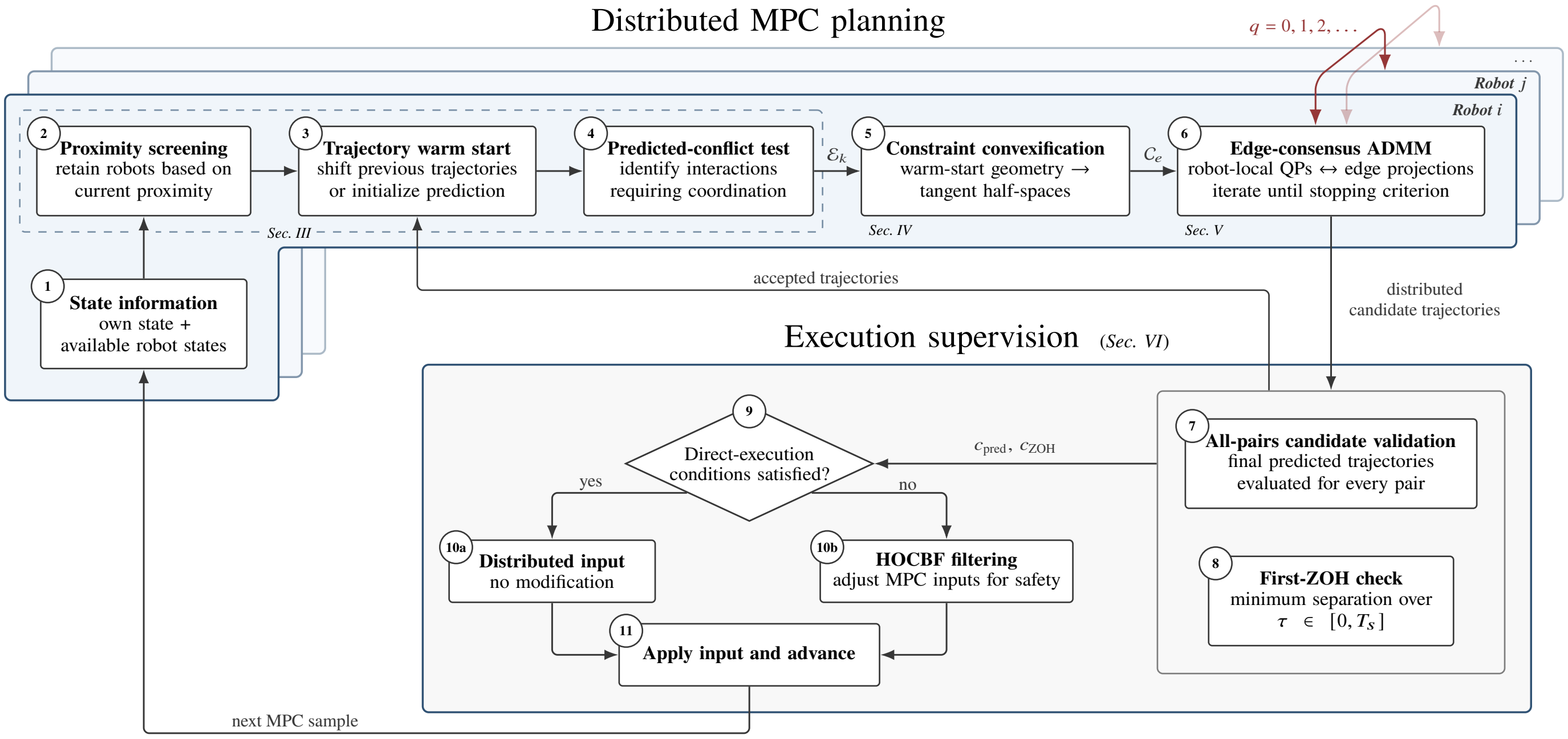}%
        }%
    }
    \caption{Overview of the proposed control framework, combining
distributed MPC planning with centralized execution supervision.}
    \label{fig:framework}
\end{figure*}

The proposed control procedure is summarized in Fig.~\ref{fig:framework}. At each sampling instant, distributed MPC planning generates candidate trajectories (Sections~III--V). A centralized execution supervisor then validates the predicted trajectories and evaluates the first control interval to determine whether the distributed inputs can be applied directly or require HOCBF filtering (Section~VI).

\section{Dynamic Predicted-Conflict Graph}

At each MPC update, robot $i$ is assumed to have access to its own current state and to the states of all robots within distance $R_{\mathrm{off}}$, although information from more distant robots may also be available. 

The pairwise couplings handled by the distributed 
solver are selected through three stages: proximity screening, trajectory
warm-start construction, and predicted-conflict testing. These stages define
the time-varying interaction graph
\begin{equation}
    \mathcal G_k=(\Ical,\Ecal_k),
\end{equation}
where $\Ecal_k$ contains the active robot pairs. The graph sparsifies the
distributed coordination problem, while candidate validation and execution
supervision are performed separately after ADMM termination.

\subsection{Proximity screening}

A preliminary selection based on inter-robot distance restricts the subsequent predicted-conflict tests to nearby robot pairs,
reducing the number of trajectory comparisons required to
construct the interaction graph.
For a robot pair $(i,j)$, define the current distance as
\begin{equation}
    d_{ij,k}=\norm{p_{i,k}-p_{j,k}}_2.
\end{equation}
Candidate pairs are selected using distance hysteresis. A pair that was
inactive at the previous MPC sample, $(i,j)\notin\Ecal_{k-1}$, becomes a
candidate when
\begin{equation}
    d_{ij,k}\leq R_{\mathrm{on}},
\end{equation}
whereas a previously active pair, $(i,j)\in\Ecal_{k-1}$, remains a candidate
while
\begin{equation}
    d_{ij,k}\leq R_{\mathrm{off}},
    \qquad R_{\mathrm{off}}>R_{\mathrm{on}}.
\end{equation}
The larger retention radius reduces repeated entry and removal of pairs
near the proximity boundary, limiting unnecessary changes in the pairwise couplings
across MPC samples.

\subsection{Trajectory warm start}
\label{subsec:warm_start}
When a previous validated candidate is available, its input sequence is
used to update the warm-start memory. Otherwise, the previous warm-start
sequence is retained. At the next MPC sample, the available sequence is
shifted by one stage and completed with a zero terminal acceleration. If
no previous sequence is available, a path-directed proportional--derivative
guess is constructed from the current state using a short look-ahead point
along the reference path and a tangent velocity reference. Let
$\bar p_{i,\ell}$ denote the resulting warm-start position of robot $i$ at
prediction stage $\ell$.

For each candidate pair $(i,j)$, let $\tilde p_{j,\ell|i}$ denote the
prediction of robot $j$ available to robot $i$ for conflict detection. If
$(i,j)\notin\Ecal_{k-1}$, a constant-velocity prediction is used,
\begin{equation}
    \tilde p_{j,\ell|i}
    =p_{j,k}+\ell T_s v_{j,k},
    \qquad \ell=1,\ldots,N.
\end{equation}
For a previously active pair $(i,j)\in\Ecal_{k-1}$,
$\tilde p_{j,\ell|i}$ is instead the shifted warm-start trajectory
previously communicated by robot $j$. Robot $j$ constructs its prediction
of robot $i$ analogously.

Besides numerical initialization, the warm-start trajectories are used in the predicted-conflict detection and determine the relative geometry used for subsequent collision-avoidance constraint convexification.

\subsection{Predicted-conflict test}

The predicted-conflict margin depends on whether the pair was inactive
or active at the previous MPC sample. The prescribed parameters
$m_{\mathrm{on},1}$ and $m_{\mathrm{on},N}$ specify the margins at the
first and final prediction stages for previously inactive pairs,
whereas $m_{\mathrm{off},1}$ and $m_{\mathrm{off},N}$ specify the
corresponding margins for previously active pairs. The endpoint
values are selected as
\begin{equation}
    (m_1,m_N)=
    \begin{cases}
        (m_{\mathrm{on},1},m_{\mathrm{on},N}),
        & (i,j)\notin\Ecal_{k-1},\\
        (m_{\mathrm{off},1},m_{\mathrm{off},N}),
        & (i,j)\in\Ecal_{k-1}.
    \end{cases}
\end{equation}
The margin is linearly interpolated along the prediction horizon as
\begin{equation}
    m_\ell
    =m_1+\frac{\ell-1}{N-1}(m_N-m_1),
    \qquad \ell=1,\ldots,N.
\end{equation}
In the proposed design, $m_N<m_1$, so near-term conflicts are screened
more conservatively, while the smaller far-horizon margin limits
unnecessary coupling.

Endpoint $i$ requests coordination with robot $j$ if
\begin{equation}\label{eq:request}
    \exists\,\ell\in\{1,\ldots,N\}:\quad
    \norm{\bar p_{i,\ell}-\tilde p_{j,\ell|i}}_2
    \leq d_{ij}^{\min}+m_\ell.
\end{equation}
Robot $j$ independently performs the reciprocal test. A request from
either endpoint activates the undirected edge $(i,j)\in\Ecal_k$.
Once activated, the endpoints exchange their warm-start trajectories
to construct the separating half-spaces before ADMM begins.

No neighbor cap is imposed: every candidate pair satisfying the
predicted-conflict test in either direction is included in $\Ecal_k$.
The proximity radii govern candidate selection, whereas the ``on''/``off''
predicted-conflict margins govern edge activation and persistence.
The active graph can therefore be constructed through local computations
and pairwise information exchange, without centralized coordination.

\section{Constraint Convexification}

The collision-avoidance constraint is
\begin{equation}\label{eq:exact_circle}
    \norm{p_{i,\ell}-p_{j,\ell}}_2\geq d_{ij}^{\min},
    \quad
    (i,j)\in\Ecal_k,\quad \ell=1,\ldots,N.
\end{equation}
Its feasible set is nonconvex. Using the warm-start relative position
\begin{equation}
    \bar r_{ij,\ell}
    =
    \bar p_{i,\ell}-\bar p_{j,\ell},
\end{equation}
the corresponding unit normal is constructed as
\begin{equation}\label{eq:normal}
    n_{ij,\ell}
    =
    \frac{\bar r_{ij,\ell}}
    {\norm{\bar r_{ij,\ell}}_2}.
\end{equation}
If the warm-start relative position is numerically degenerate, the normal is instead computed from the current measured relative position.

For each prediction stage $\ell=1,\ldots,N$, the circular exclusion
constraint~\eqref{eq:exact_circle} is approximated by the
stage-dependent tangent half-space
\begin{equation}\label{eq:tangent_plane}
    n_{ij,\ell}\trans(p_{i,\ell}-p_{j,\ell})
    \geq d_{ij}^{\min}.
\end{equation}

Since $\norm{n_{ij,\ell}}_2=1$, satisfaction of
\eqref{eq:tangent_plane} at each prediction stage implies the corresponding
sampled separation constraint in~\eqref{eq:exact_circle}.
To collect the stagewise collision constraints, define the predicted position
sequence of robot $i$ as
\begin{equation}
    P_i=
    \begin{bmatrix}
        p_{i,1} & \cdots & p_{i,N}
    \end{bmatrix}
    \in\R^{2\times N}.
    \label{eq:position_sequence}
\end{equation}
For each active edge $e=(i,j)\in\Ecal_k$, define the convex pairwise feasible set
\begin{multline}
    \mathcal C_e=
    \left\{
    (P_i,P_j):
    n_{ij,\ell}\trans
    \left(p_{i,\ell}-p_{j,\ell}\right)
    \geq d_{ij}^{\min},
    \right.\\
    \left.
    \ell=1,\ldots,N
    \right\}.
    \label{eq:pairwise_convex_set}
\end{multline}

Only one set of collision-plane normals is constructed per MPC sample, and the planes therefore remain fixed throughout the subsequent ADMM iterations.
Since the remaining dynamics and constraints are linear and the objective is quadratic, this single convexification yields a fixed convex QP. 
The tangent half-spaces preserve the local avoidance side induced by the
warm-start trajectories. Consequently, the solution of the convexified problem
depends on the warm-start geometry and may differ for warm starts corresponding
to different avoidance maneuvers. 

\begin{figure*}[t]
    \centering
    \includegraphics[width=\textwidth]
        {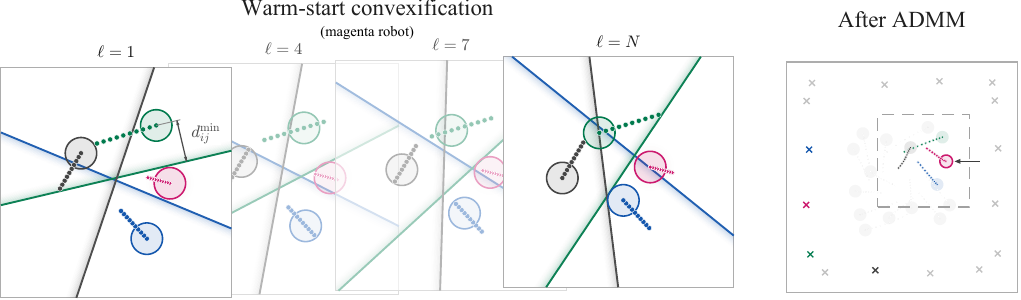}
    \caption{Illustration of collision-avoidance convexification.
    Left: warm-start trajectories (dots), predicted positions (disks),
    and separating half-spaces for the magenta robot along the horizon;
    shading indicates the forbidden sides.
    Right: current fleet configuration, updated predicted trajectories after ADMM, and goals (crosses). The dashed box identifies the enlarged region.}
    \label{fig:plane_R7_k20}
\end{figure*}

Figure~\ref{fig:plane_R7_k20} illustrates the convexification procedure
in a simulated crossing scenario involving 16 robots. Focusing on
the magenta robot, it shows how the warm-start trajectories determine
the separating half-spaces along the prediction horizon.

\section{Edge-Consensus ADMM}

\subsection{Consensus reformulation}

For each active edge $e=(i,j)\in\Ecal_k$, the convexified collision-avoidance coupling $(P_i,P_j)\in\mathcal C_e$ prevents a direct decomposition of the MPC problem by robot. To separate this coupling, introduce edge-local copies  
$Z_i^e,Z_j^e\in\R^{2\times N}$ of the corresponding owner trajectories. Let
$z_{i,\ell}^e$ and $z_{j,\ell}^e$ denote their stage-$\ell$ columns, and define the 
edge-feasible set
\begin{multline}\label{eq:edge_set}
    \mathcal H_e=
    \left\{
    (Z_i^e,Z_j^e):
    n_{ij,\ell}\trans
    \left(z_{i,\ell}^e-z_{j,\ell}^e\right)
    \geq d_{ij}^{\min}, \right. \\
    \left. \ell=1,\ldots,N
    \right\}.
\end{multline}

Using the robot-local feasible sets $\mathcal F_i$ defined in Section~II, the convexified multi-robot optimization problem related to the MPC strategy can then be written in consensus form as
\begin{align}
    \min_{\{\xi_i,u_i\},\{Z^e\}} 
    \quad
    &\sum_{i\in\Ical}J_i(\xi_i,u_i)
    +\sum_{e\in\Ecal_k}
    I_{\mathcal H_e}\!\left(Z_i^e,Z_j^e\right)
    \label{eq:consensus_problem}\\
    \text{s.t.}\quad
    &(\xi_i,u_i)\in\mathcal F_i,
    \qquad i\in\Ical,
    \nonumber\\
    &P_i=Z_i^e,\qquad
    P_j=Z_j^e,
    \qquad e=(i,j)\in\Ecal_k,
    \nonumber
\end{align}
where $I_{\mathcal H_e}$ is zero when its arguments belong to
$\mathcal H_e$ and $+\infty$ otherwise. Since the active graph and separating normals $n_{ij,\ell}$ remain fixed throughout the ADMM inner loop, \eqref{eq:consensus_problem} is a fixed convex QP at each MPC sample. The convergence properties of the resulting ADMM scheme and the assumptions under which they hold are discussed in Section~\ref{sec:admm-convergence}.

The formulation in \eqref{eq:consensus_problem} separates the problem into two blocks with complementary decompositions. The robot block decomposes over $i$, with each robot handling its own dynamics, local constraints, and objective. The edge block decomposes over $e\in\Ecal_k$ and, within each edge, over the prediction stages because the half-space constraints in~\eqref{eq:edge_set} are stagewise independent. ADMM alternates between these two blocks: the robots solve independent local
QPs, while the active edges perform independent half-space projections.

\subsection{Robot-local QP update}
Let
\begin{equation}
    \Ncal_i
    =
    \left\{e\in\Ecal_k:\text{$e$ is incident to robot $i$}\right\}
\end{equation}
denote the set of active edges incident to robot $i$. For each
$e\in\Ncal_i$, let $W_i^e\in\R^{2\times N}$ be the scaled dual variable
associated with the consensus constraint $P_i=Z_i^e$.

At ADMM round $q+1$, robot $i$ solves
\begin{align}
    \min_{\xi_i,u_i}\quad
    &J_i(\xi_i,u_i)
    +\frac{\rho}{2}
    \sum_{e\in\Ncal_i}
    \norm{P_i-Z_i^{e,q}+W_i^{e,q}}_F^2
    \label{eq:node_update}\\
    \text{s.t.}\quad
    &(\xi_i,u_i)\in\mathcal F_i,
    \nonumber
\end{align}
where $\rho>0$ is the ADMM penalty parameter. During this update,
$Z_i^{e,q}$ and $W_i^{e,q}$ are fixed. Hence, no pairwise
collision-avoidance constraint appears explicitly in the robot-local QP;
interaction with neighboring robots enters only through the quadratic
consensus penalties.

The resulting robot-local problems are convex QPs and are solved with
HPIPM~\cite{frison2020}. Since the edge copies and dual variables are fixed
during this phase, all robot-local QPs can be solved in parallel.

\subsection{Edge projection and dual update}

To accelerate consensus, the implementation uses ADMM over-relaxation. After
the robot-local update, the owner trajectory associated with edge $e=(i,j)$ is
over-relaxed as
\begin{equation}\label{eq:admm_relaxation}
    \widehat P_i^{e,q+1}
    =
    \alpha P_i^{q+1}
    +(1-\alpha)Z_i^{e,q},
    \qquad \alpha\in(0,2),
\end{equation}
with an analogous expression for endpoint $j$.

For each prediction stage $\ell=1,\ldots,N$, define the dual-shifted
over-relaxed positions
\begin{equation}
    y_{i,\ell}
    =
    \widehat p_{i,\ell}^{e,q+1}+w_{i,\ell}^{e,q},
    \qquad
    y_{j,\ell}
    =
    \widehat p_{j,\ell}^{e,q+1}+w_{j,\ell}^{e,q},
\end{equation}
where $\widehat p_{i,\ell}^{e,q+1}$ and $w_{i,\ell}^{e,q}$ denote the
stage-$\ell$ columns of $\widehat P_i^{e,q+1}$ and $W_i^{e,q}$,
respectively.

The edge update admits an exact solution and therefore requires no iterative constrained optimization.
For a scalar $s$, let $[s]_+=\max\{s,0\}$.

\begin{lemma}[Exact projection]
\label{lem:jbb-projection}
Fix an edge $e=(i,j)$ and a prediction stage $\ell$. Let $\|n_{ij,\ell}\|_2=1$ and let $b\in\mathbb R$. The problem
\begin{equation}
\begin{aligned}
\min_{z_i,z_j\in\mathbb R^2}\quad &
 \tfrac12\|z_i-y_{i,\ell}\|_2^2+
 \tfrac12\|z_j-y_{j,\ell}\|_2^2\\
\text{subject to}\quad &n_{ij,\ell}\trans(z_i-z_j)\ge b
\end{aligned}
\label{eq:jbb-projection-problem}
\end{equation}
has the unique solution
\begin{equation}
\begin{aligned}
z_i^\star&=y_{i,\ell}+\tfrac12[\delta_b]_+n_{ij,\ell},\\
z_j^\star&=y_{j,\ell}-\tfrac12[\delta_b]_+n_{ij,\ell},\\
\delta_b&=b-n_{ij,\ell}\trans(y_{i,\ell}-y_{j,\ell}).
\end{aligned}
\label{eq:jbb-projection-solution}
\end{equation}
Here, $(z_i^\star,z_j^\star)$ denotes the unique minimizer of the projection problem. 
\end{lemma}

\begin{proof}
See Appendix~\ref{app:proof_lemma1}.
\end{proof}

Taking $b=d_{ij}^{\min}$ in Lemma~1 gives the stagewise ADMM edge update
\[
z_{i,\ell}^{e,q+1}=z_i^\star,
\qquad
z_{j,\ell}^{e,q+1}=z_j^\star.
\]
Since the edge-feasible set is a Cartesian product of stagewise half-spaces, these projections can be evaluated independently for every prediction stage and active edge.
The projected columns form $Z_i^{e,q+1}$ and $Z_j^{e,q+1}$.

After all stagewise projections, the scaled dual variables are updated as
\begin{equation}\label{eq:dual_update}
    W_i^{e,q+1}
    =
    W_i^{e,q}
    +
    \widehat P_i^{e,q+1}
    -
    Z_i^{e,q+1},
\end{equation}
with an analogous update for endpoint $j$.

\subsection{ADMM convergence for the fixed convexified problem}
\label{sec:admm-convergence}

The robot updates in \eqref{eq:node_update} form one separable ADMM block, while the projections onto \eqref{eq:edge_set} form the second block. Together with the over-relaxation in \eqref{eq:admm_relaxation} and the dual update in \eqref{eq:dual_update}, they define a relaxed two-block ADMM scheme for \eqref{eq:consensus_problem}. The following proposition establishes its convergence at a fixed MPC sample.

For the convergence analysis, define the primal residual at ADMM iteration $q$ as
\begin{equation}
r_p^q =
\left[
\sum_{e=(i,j)\in\mathcal E_k}
\left(
\|P_i^q-Z_i^{e,q}\|_F^2
+
\|P_j^q-Z_j^{e,q}\|_F^2
\right)
\right]^{1/2}.
\label{eq:admm-primal-residual}
\end{equation}
It measures the disagreement between the robot-owned trajectories and their edge-local copies.

\begin{proposition}[Convergence at a fixed MPC sample]
\label{prop:jbb-convergence}
Fix the current augmented initial conditions, the warm-start data, the active edge set $\mathcal E_k$ and the unit normals $n_{ij,\ell}$ throughout the ADMM loop. Suppose that
\begin{enumerate}
\item the local feasible sets $\mathcal F_i$ in \eqref{eq:consensus_problem} are nonempty compact convex polyhedra, and the objectives $J_i$ in \eqref{eq:local_cost} are continuous convex quadratics;
\item the unaugmented Lagrangian of \eqref{eq:consensus_problem} admits a primal-dual saddle point;
\item each minimization in \eqref{eq:node_update} and each edge projection is solved exactly, with fixed $\rho>0$ and $\alpha\in(0,2)$.
\end{enumerate}
Let $(\xi_i^q,u_i^q)$ denote the robot-owned sequences in \eqref{eq:state_input} at iteration $q$, and let $J_k^\star$ be the optimal value of \eqref{eq:consensus_problem}. If the iterations are continued without activating either stopping branch or the finite cap, the primal residual $r_p^q$ defined in \eqref{eq:admm-primal-residual} satisfies $r_p^q\to0$, and
\begin{equation}
\sum_{i\in\mathcal I}J_i(\xi_i^q,u_i^q)\longrightarrow J_k^\star.
\label{eq:jbb-objective-convergence}
\end{equation}
Every accumulation point of the robot-owned sequences is an optimizer of \eqref{eq:consensus_problem}.

Since \(w_0>0\) in \eqref{eq:prox_weight}, all optimizers of \eqref{eq:consensus_problem} share the same stacked acceleration sequence, and the iterates converge to that sequence. In particular,
\begin{equation}
\max_{i\in\mathcal I}\|a_{i,0}^{q}-a_{i,0}^{q-1}\|_2\longrightarrow0.
\label{eq:jbb-control-increment}
\end{equation}
\end{proposition}

\begin{proof}
See Appendix~\ref{app:proof_proposition1}.
\end{proof}

With fixed augmented initial conditions, finite input bounds and finite horizon, the local dynamics bound the predicted states, consistently with the compactness assumption. The weight assumptions in Section~\ref{sec:localMPCobj} ensure convexity
of \eqref{eq:local_cost}, while $w_0>0$ and $0<\beta<1$ imply
$w_\ell>0$ at every input stage, yielding strict convexity
with respect to the stacked acceleration sequence. A unit normal must be assigned explicitly when a warm-start relative position vanishes. Local feasibility alone does not establish the saddle-point assumption for \eqref{eq:consensus_problem}. No connectivity assumption is needed: disconnected components decouple. If $\mathcal E_k=\varnothing$, \eqref{eq:consensus_problem} reduces to independent local minimizations and no consensus loop is required.

Proposition~\ref{prop:jbb-convergence} establishes asymptotic convergence for \eqref{eq:consensus_problem} at one MPC sample. It does not establish recursive feasibility across samples or guarantee convergence within $q_{\max}$ rounds. This fact motivates a certificate that can be evaluated at a finite iterate.

\subsection{ADMM termination criteria}

ADMM uses two early-stopping branches based on residual convergence and a finite-iteration feasibility certificate; otherwise, the iterations terminate upon reaching the maximum round budget $q_{\max}$.

The residual-based stopping criterion uses the primal residual in \eqref{eq:admm-primal-residual} and the dual residual defined as follows:
\begin{equation}\label{eq:dual_res}
r_{\mathrm d}^{q+1}
=
\rho
\left[
\sum_{i=1}^{M}
\left\|
\sum_{e\in\Ncal_i}
\left(Z_i^{e,q+1}-Z_i^{e,q}\right)
\right\|_F^2
\right]^{1/2}.
\end{equation}
After a prescribed minimum number of rounds, the residual branch terminates
when the primal and dual residuals satisfy the standard dimension-scaled
absolute and relative tolerances~\cite{boyd2011}.

A second stopping branch allows termination before full owner--edge consensus
when the robot-owned trajectories are already certified to satisfy the active
separating half-spaces. For every active edge $e=(i,j)$ and prediction stage
$\ell$, define
\begin{multline}\label{eq:finite_round_certificate}
    \eta_{e,\ell}^{q}
    ={}n_{ij,\ell}\trans
    \left(z_{i,\ell}^{e,q}-z_{j,\ell}^{e,q}\right)-d_{ij}^{\min}\\
    -\norm{p_{i,\ell}^{q}-z_{i,\ell}^{e,q}}_2
    -\norm{p_{j,\ell}^{q}-z_{j,\ell}^{e,q}}_2.
\end{multline}

The worst-case certificate over all active edges and prediction stages is
\begin{equation}
    \label{eq:worst_case_certif}
    \eta_{\min}^{q}
    =
    \min_{e\in\Ecal_k}
    \min_{\ell=1,\ldots,N}
    \eta_{e,\ell}^{q}.
\end{equation}
When $\mathcal E_k=\varnothing$, the ADMM loop and certificate evaluation are skipped, and the robot-local QPs are solved independently.
The quantity in \eqref{eq:finite_round_certificate} bounds the separation of the robot-owned positions, rather than only that of the edge-local copies. Lemma \ref{lem:jbb-separation} establishes this interpretation and relates the bound to the primal residual in \eqref{eq:admm-primal-residual}. 

\begin{lemma}[Separation bound under incomplete consensus]
\label{lem:jbb-separation}
Fix an MPC sample with $\mathcal E_k\ne\varnothing$, and assume $\|n_{ij,\ell}\|_2=1$ for every active edge and prediction stage. At a common iteration $q$, let $p_{i,\ell}^q$ be the stage-$\ell$ column of the robot-owned trajectory $P_i^q$, and let $z_{i,\ell}^{e,q}$ be the corresponding column of the edge-local copy $Z_i^{e,q}$. For $e=(i,j)\in\mathcal E_k$ and $\ell\in\{1,\ldots,N\}$, define the signed edge-copy margin
\begin{equation}
s_{e,\ell}^q=
 n_{ij,\ell}\trans(z_{i,\ell}^{e,q}-z_{j,\ell}^{e,q})-d_{ij}^{\min}.
\label{eq:jbb-copy-margin}
\end{equation}
With $\eta_{e,\ell}^q$ and $\eta_{\min}^q$ as defined in \eqref{eq:finite_round_certificate} and \eqref{eq:worst_case_certif}, respectively, the following inequalities hold:
\begin{equation}
\begin{aligned}
\|p_{i,\ell}^q-p_{j,\ell}^q\|_2-d_{ij}^{\min}
&\ge n_{ij,\ell}\trans(p_{i,\ell}^q-p_{j,\ell}^q)-d_{ij}^{\min}\\
&\ge\eta_{e,\ell}^q\\
&\ge s_{e,\ell}^q-\sqrt{2}\,r_p^q.
\end{aligned}
\label{eq:jbb-edge-bound}
\end{equation}
Consequently,
\begin{equation}
\begin{aligned}
&\min_{\substack{e=(i,j)\in\mathcal E_k\\1\le\ell\le N}}
 \bigl(\|p_{i,\ell}^q-p_{j,\ell}^q\|_2-d_{ij}^{\min}\bigr)
 \ge\eta_{\min}^q,\\
&\eta_{\min}^q\ge
 \min_{\substack{e\in\mathcal E_k\\1\le\ell\le N}}s_{e,\ell}^q
 -\sqrt{2}\,r_p^q.
\end{aligned}
\label{eq:jbb-global-bound}
\end{equation}
In particular, $\eta_{\min}^q\ge\eta_{\mathrm{cert}}\ge0$ implies
\begin{equation}
\begin{aligned}
n_{ij,\ell}\trans(p_{i,\ell}^q-p_{j,\ell}^q)
 &\ge d_{ij}^{\min}+\eta_{\mathrm{cert}},\\
\|p_{i,\ell}^q-p_{j,\ell}^q\|_2
 &\ge d_{ij}^{\min}+\eta_{\mathrm{cert}}
\end{aligned}
\label{eq:jbb-certified-owner-separation}
\end{equation}
for all active edges and prediction stages.
\end{lemma}

\begin{proof}
See Appendix~\ref{app:proof_lemma2}.
\end{proof}

Lemma~\ref{lem:jbb-separation} shows that the certificate in \eqref{eq:worst_case_certif} can certify robot-owned half-space constraint satisfaction in \eqref{eq:tangent_plane} before consensus is reached. The bound is geometric and does not require ADMM convergence or feasibility of edge copies; after an exact projection onto \eqref{eq:edge_set}, the additional property $s_{e,\ell}^q\ge0$ holds. However, a small positive residual alone need not certify nonnegative clearance, since the edge-copy margins may be zero. The certificate concerns only active pairs at prediction stages, so all-pairs and intersample validation remains necessary. This motivates using the certificate together with a first-input stabilization test in the stopping criterion below.

After its prescribed minimum number of rounds, the certificate branch
terminates when
\begin{subequations}\label{eq:certificate_stop_conditions}
\begin{align}
    \eta_{\min}^{q} &\geq \eta_{\mathrm{cert}},\\
    \max_i
    \norm{a_{i,0}^{q}-a_{i,0}^{q-1}}_2 &\leq \epsilon_a.
\end{align}
\end{subequations}

The second condition additionally requires the first control action to have stabilized between consecutive ADMM rounds.

If neither stopping branch is satisfied, the iterations continue until the
maximum number of rounds $q_{\max}$ is reached.

\section{Execution Supervision}
\label{sec:execution_safety}

\subsection{Candidate validation and first-ZOH check}

After the ADMM inner loop terminates, the final robot trajectories are
evaluated over all robot pairs, including pairs that are not active in
$\Ecal_k$. Their minimum clearance is
\begin{equation}\label{eq:pred_clearance}
    c_{\mathrm{pred}}
    =
    \min_{i<j}\min_{\ell=1,\ldots,N}
    \left(
    \norm{p_{i,\ell}-p_{j,\ell}}_2-d_{ij}^{\min}
    \right).
\end{equation}
The candidate solution is validated when the robot-local QPs have been solved successfully and $c_{\mathrm{pred}}\geq 0$. This validation is performed independently of the ADMM stopping reason, so a candidate obtained upon reaching $q_{\max}$ may still be accepted. A numerically valid candidate that fails validation is not executed directly,
but its first input is retained as the reference for the HOCBF filter.

Since satisfaction of the collision constraints at the prediction stages does
not exclude intersample collisions, the first control input is additionally
checked over the complete zero-order-hold interval. The exact first-interval
clearance is defined as
\begin{equation}
c_{\mathrm{ZOH}}
=
\min_{i<j}\min_{\tau\in[0,T_s]}
\left(d_{ij}(\tau)-d_{ij}^{\min}\right),
\end{equation}
where $d_{ij}(\tau)$ denotes the pairwise distance under the corresponding
constant-acceleration relative motion. 

\subsection{HOCBF filtering}

When required, the execution filter minimally modifies the distributed
reference accelerations subject to pairwise HOCBF constraints. For each robot
pair, let
\begin{equation}
    r_{ij}=p_{i,k}-p_{j,k},
    \qquad
    v_{ij}=v_{i,k}-v_{j,k}
\end{equation}
denote the current relative position and velocity, respectively, and define
\begin{equation}\label{eq:hocbf_h}
    h_{ij}
    =
    r_{ij}\trans r_{ij}
    -
    \left(d_{ij}^{\min}\right)^2.
\end{equation}
For the double-integrator dynamics, the relative-degree-two HOCBF condition is
\begin{equation}\label{eq:hocbf_row}
    2r_{ij}\trans(a_i-a_j)+\gamma_{ij}\geq0,
\end{equation}
where $
    \gamma_{ij}
    =
    2v_{ij}\trans v_{ij}
    +2(\alpha_1+\alpha_2)r_{ij}\trans v_{ij}
    +\alpha_1\alpha_2 h_{ij}$,
and $\alpha_1,\alpha_2>0$ are the HOCBF gains~\cite{xiao2022,ames2019}.

Only robot pairs whose current separation does not exceed a prescribed activation distance $d_{\mathrm{act}}$ are included in the filter. For the homogeneous robot fleet
considered in the simulations, this distance is selected using the heuristic rule
\begin{equation}
d_{\mathrm{act}}
=
d_{ij}^{\min}
+\frac{(v_i^{\max})^2}{a_i^{\max}}
+2v_i^{\max}T_s
+d_{\mathrm{buf}},
\label{eq:hocbf-activation-distance}
\end{equation}
where $d_{\mathrm{buf}}\ge0$ is an additional activation buffer. The second and third terms are motivated by the combined stopping distance and the relative displacement over one sampling interval, respectively, for two identical robots approaching head-on along a coordinate axis at
speed $v_i^{\max}$, with braking acceleration magnitude $a_i^{\max}$.

Since the bounds in \eqref{eq:box_limits} are componentwise, this one-dimensional construction is not a general stopping-distance bound for arbitrary planar encounters. Accordingly, \eqref{eq:hocbf-activation-distance} is used to select the filter activation neighborhood, rather than as a sufficient
condition for collision avoidance.

The centralized HOCBF-QP is
\begin{align}
    \min_{a_1,\ldots,a_M}\quad
    &\frac{1}{2}\sum_{i=1}^{M}
    \|a_i-a_i^{\mathrm{ref}}\|_2^2
    \label{eq:hocbf_qp}\\
    \text{s.t.}\quad
    &\eqref{eq:hocbf_row}
    \text{ for all active filter pairs,}
    \nonumber\\
    &-a_i^{\mathrm{peak}} \mathds{1}
    \preceq a_i
    \preceq a_i^{\mathrm{peak}}\mathds{1},
    \nonumber\\
    &-v_i^{\max}\mathds{1}
    \preceq v_{i,k}+T_s a_i
    \preceq v_i^{\max}\mathds{1},
    \nonumber
\end{align}
where $\mathds{1}$ is the all-ones vector, $a_i^{\mathrm{ref}}$ denotes the first candidate MPC acceleration, and
$a_i^{\mathrm{peak}}>a_i^{\max}$ allows larger execution-level corrective
accelerations during safety intervention.

The filter is bypassed when the distributed candidate is validated and $c_{\mathrm{ZOH}}\geq 0$.
Otherwise, the HOCBF-QP is solved. The first-interval clearance check is performed on the unfiltered candidate. The filtered input is applied without an additional ZOH clearance acceptance test; therefore, feasibility of \eqref{eq:hocbf_qp} alone does not certify continuous-time separation over the sampling interval. If the QP fails, a bounded braking input opposing the current velocity is applied as a fallback.

Algorithm~\ref{alg:closed_loop} summarizes the complete MPC procedure.

\begin{algorithm}[t]
\caption{Distributed MPC with selective HOCBF filtering}
\label{alg:closed_loop}
\begin{algorithmic}[1]
\State Measure the robot states and update the trajectory warm starts.
\State Build $\Ecal_k$, exchange warm-start trajectories over active edges,
and construct the separating half-spaces.
\State Initialize the ADMM edge variables.
\For{$q=0,\ldots,q_{\max}-1$}
    \State Solve the robot-local QPs in parallel.
    \State Exchange updated trajectories over active edges.
    \State Over-relax, project the edge copies, and update the scaled dual variables.
    \If{the residual- or certificate-based stopping criterion is satisfied}
        \State \textbf{break}
    \EndIf
\EndFor
\State Extract the distributed candidate and validate its predicted trajectories over all robot pairs.
\State Evaluate the exact all-pairs clearance over the first ZOH interval.
\If{the distributed candidate satisfies the bypass conditions}
    \State Apply the distributed input.
\Else
    \State Solve and apply the HOCBF-filtered input, or bounded braking if filtering fails.
\EndIf
\end{algorithmic}
\end{algorithm}

\section{Simulation Results}
\label{sec:results}

\subsection{Benchmark design and timing protocol}

The proposed architecture is evaluated in dense crossing scenarios with
$M\in\{8,16,32\}$ robots. The corresponding square workspaces have side
lengths $8$, $8\sqrt{2}$, and $16$~m, respectively, so that the workspace area scales proportionally with fleet size. Robots start near the workspace boundary and are assigned goals at roughly opposite boundary locations, producing a common central crossing. For each fleet size, $20$ randomized realizations of a common nominal crossing scenario are evaluated. Each realization is generated by perturbing the nominal start and goal locations while preserving the overall crossing structure. The same realization is evaluated with the proposed distributed controller and the centralized baseline.

For a controlled comparison, the centralized baseline solves the convexified MPC problem using identical robot dynamics,
objective function, prediction horizon, warm-start construction, tangent-plane
collision convexification, and physical constraints. The difference is that
the centralized formulation includes the convexified collision constraints for
all $M(M-1)/2$ robot pairs, whereas the distributed formulation retains only
the active edges in $\Ecal_k$ and solves the resulting problem through
edge-consensus ADMM. The same all-pairs candidate validation and selective
HOCBF execution filter are applied to both architectures. The comparison evaluates the combined architectural changes and does not isolate the individual contributions of graph sparsification, finite-iteration optimization, and parallel execution.

Table~\ref{tab:parameters} summarizes the simulation parameters, which were selected through preliminary tuning and kept fixed across all fleet sizes and randomized realizations.

\begin{table}[t]
\centering
\caption{Controller and numerical parameters used in the simulations.}
\label{tab:parameters}
\footnotesize
\setlength{\tabcolsep}{0.5pt}
\renewcommand{\arraystretch}{1.02}

\begin{tabular}{@{}p{0.4\columnwidth}cc@{}}
\toprule
\textbf{Parameter} & \textbf{Symbol} & \textbf{Value} \\
\midrule

\multicolumn{3}{@{}l}{\textbf{\textit{MPC model and objective}}} \\[1pt]
\quad Sampling time
& $T_s$
& $0.25$ s \\

\quad Prediction horizon
& $N$
& $10$ \\

\quad Robot radius
& $r_i$
& $0.30$ m \\

\quad Safety margin
& $d_{\mathrm{marg}}$
& $0.15$ m \\

\quad Velocity bound
& $v_i^{\max}$
& $0.85$ m/s \\

\quad Acceleration bound
& $a_i^{\max}$
& $0.65$ m/s$^2$ \\

\quad Contouring weight
& $q_{\mathrm c}$
& $6.0$ \\

\quad Longitudinal weight
& $q_{\mathrm l}$
& $1.0$ \\

\quad Terminal progress weight
& $q_\phi$
& $10.0$ \\

\quad Terminal cost matrix
& $P_f$
& $\operatorname{diag}(25,25,3,3)$ \\

\quad Acceleration weight
& $R_a$
& $0.08 I_2$ \\

\quad Input proximal weight
& $w_0$
& $50.0$ \\

\quad Proximal decay
& $\beta$
& $0.95$ \\

\addlinespace[3pt]
\multicolumn{3}{@{}l}{\textbf{\textit{Predicted-conflict graph}}} \\[1pt]

\quad Graph radii
& $(R_{\mathrm{on}},R_{\mathrm{off}})$
& $(4.5,\,5.0)$ m \\

\quad New-edge margins
& $(m_{\mathrm{on},1},m_{\mathrm{on},N})$
& $(1.0,\,0.4)$ m \\

\quad Persistent-edge margins
& $(m_{\mathrm{off},1},m_{\mathrm{off},N})$
& $(1.5,\,0.6)$ m \\

\addlinespace[3pt]
\multicolumn{3}{@{}l}{\textbf{\textit{ADMM and termination}}} \\[1pt]

\quad ADMM penalty
& $\rho$
& $100$ \\

\quad Over-relaxation
& $\alpha$
& $1.8$ \\

\quad Minimum ADMM rounds
& $q_{\min}$
& $3$ \\

\quad Maximum ADMM rounds
& $q_{\max}$
& $15$ \\

\quad Residual tolerances
& $(\epsilon_{\mathrm{abs}},\epsilon_{\mathrm{rel}})$
& $(5,\,5)\times 10^{-4}$ \\

\quad Certificate thresholds
& $(\eta_{\mathrm{cert}},\epsilon_a)$
& $(0.005~\mathrm{m},\,0.05~\mathrm{m/s^2})$ \\

\addlinespace[3pt]
\multicolumn{3}{@{}l}{\textbf{\textit{Execution supervision}}} \\[1pt]

\quad HOCBF gains
& $(\alpha_1,\alpha_2)$
& $(0.55,\,1.20)$ s$^{-1}$ \\

\quad HOCBF activation distance
& $d_{\mathrm{act}}$
& $2.69$ m \\

\quad HOCBF activation buffer
& $d_{\mathrm{buf}}$
& $0.40$ m \\

\quad Filter acceleration bound
& $a_i^{\mathrm{peak}}$
& $1.25$ m/s$^2$ \\

\bottomrule
\end{tabular}
\end{table}

All reported timings are obtained from the native C++ implementation using HPIPM on an AMD Ryzen~7~5700U with 16~GB RAM. All QP solves are executed single-threaded. For the distributed controller, robot-local QPs are timed sequentially for isolation, and each synchronous ADMM round is assigned the maximum local solve time across robots, consistent with the slowest-worker limitation of synchronous ADMM~\cite{chang2016asyncADMM}. Physical communication and synchronization overheads are excluded. One warm-up run is discarded per architecture and fleet size, after which each of the $20$ realizations is executed once per architecture with alternating execution order across scenario seeds.

For each scenario, the mean and empirical 90th-percentile MPC computation times are evaluated over the closed-loop updates and then averaged across the $20$ randomized realizations; the reported worst time is the largest individual MPC update observed over all scenarios. The MPC computation time includes the complete online control pipeline, including interaction-graph construction, optimization, candidate validation, and HOCBF filtering when invoked.

\begin{figure*}[t]
\centering

\subfloat{%
    \includegraphics[width=0.26725\textwidth]
    {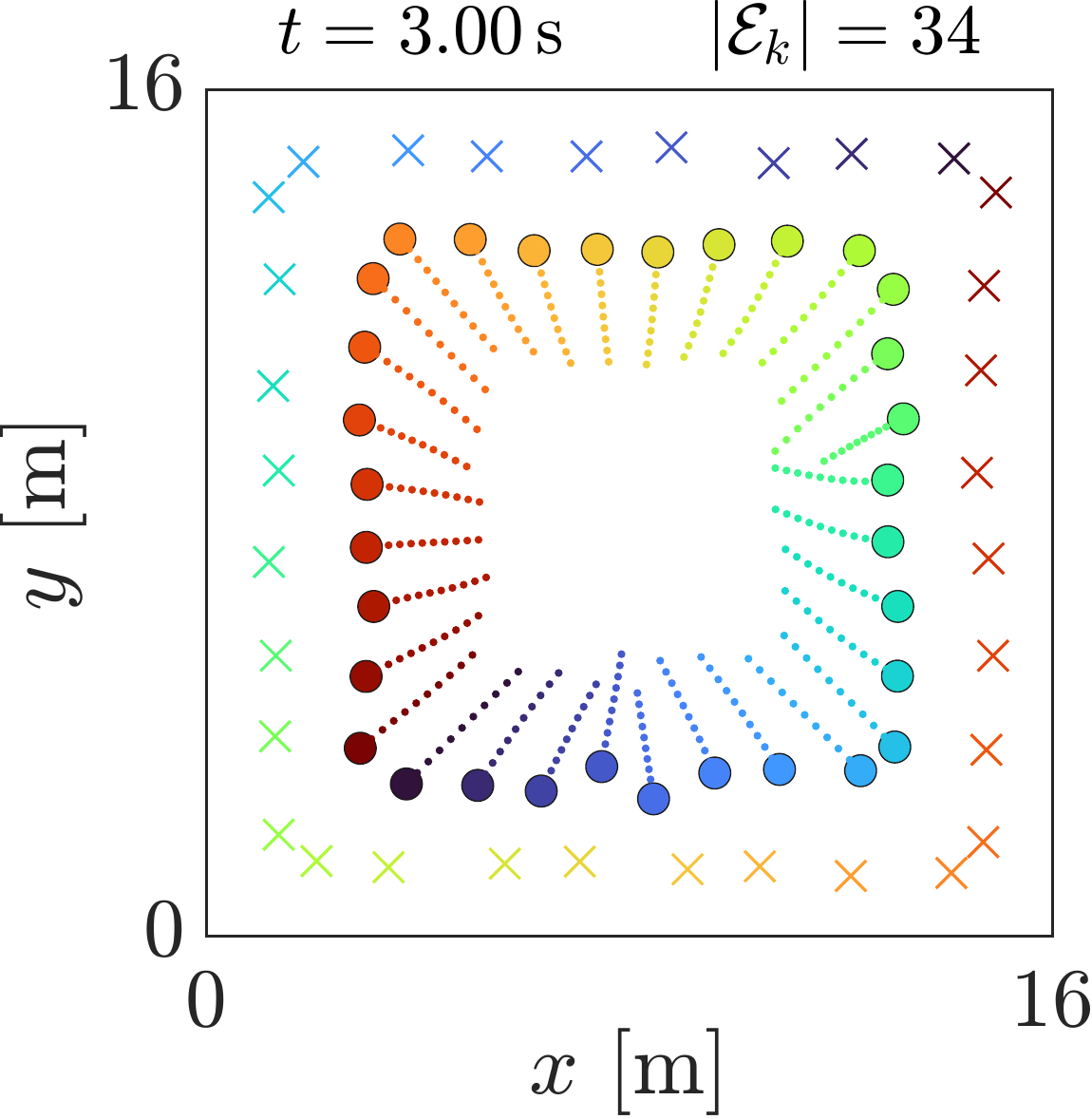}}
\hfill
\subfloat{%
    \includegraphics[width=0.24\textwidth]
    {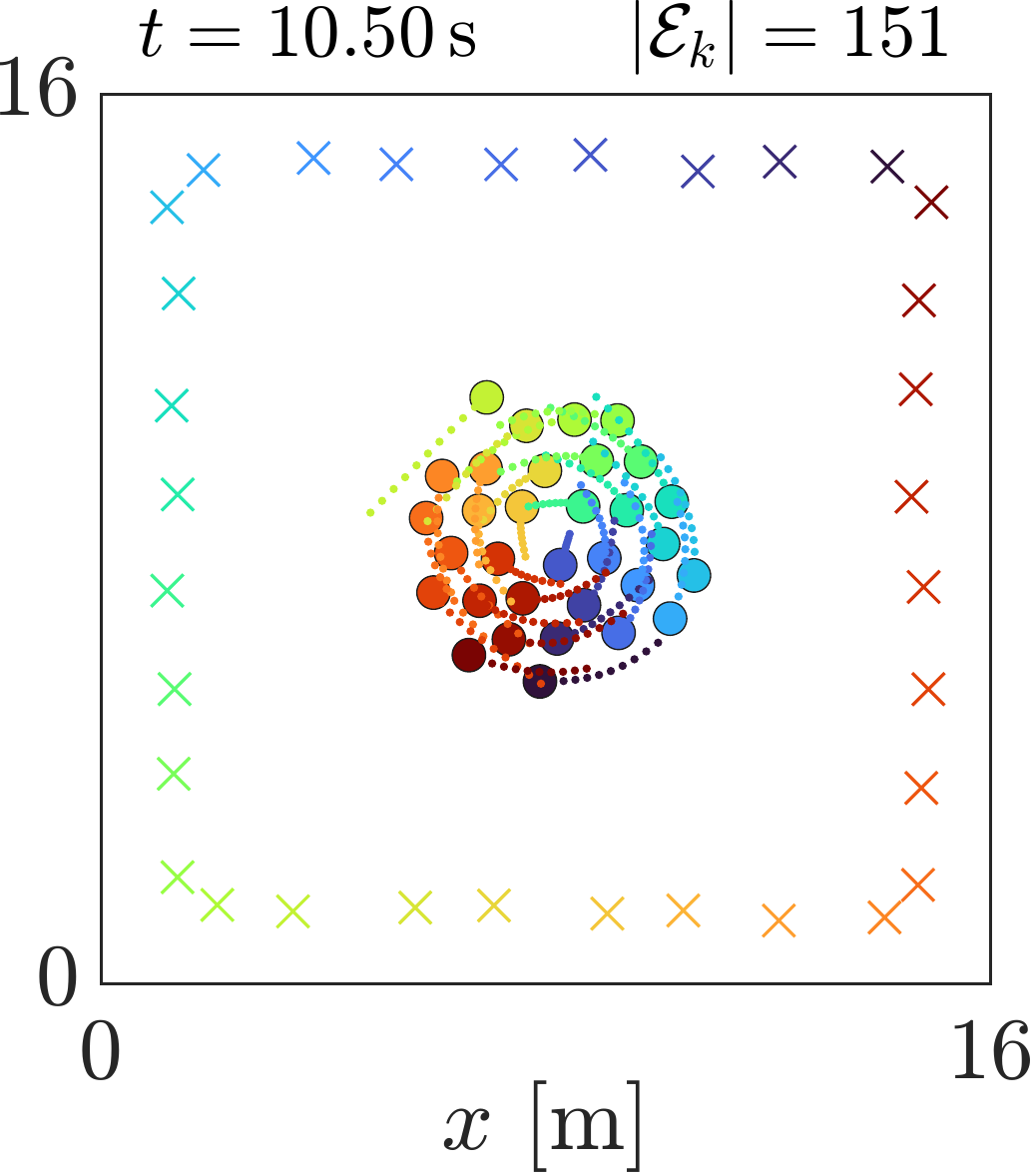}}
\hfill
\subfloat{%
    \includegraphics[width=0.24\textwidth]
    {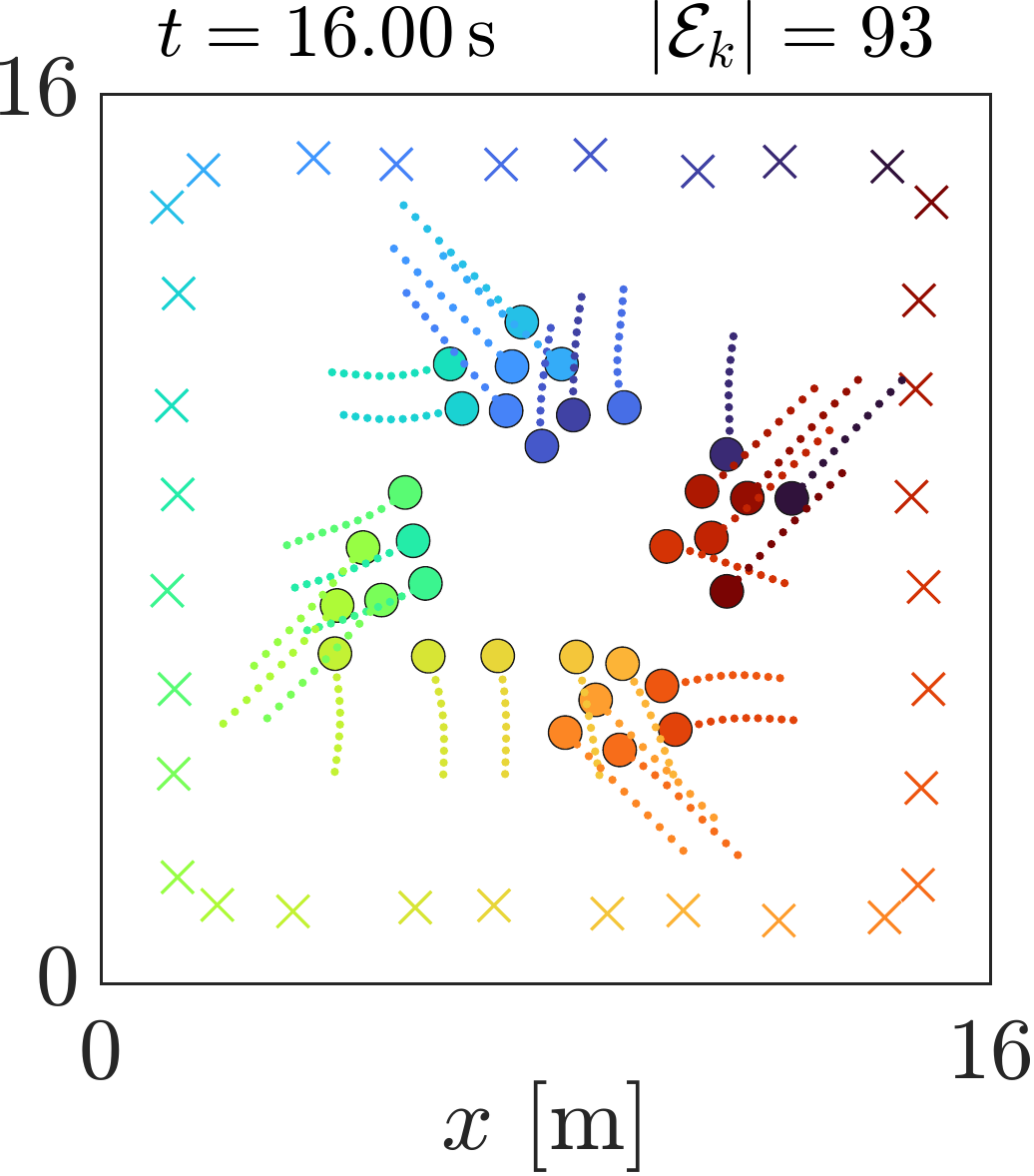}}
\hfill
\subfloat{%
    \includegraphics[width=0.24\textwidth]
    {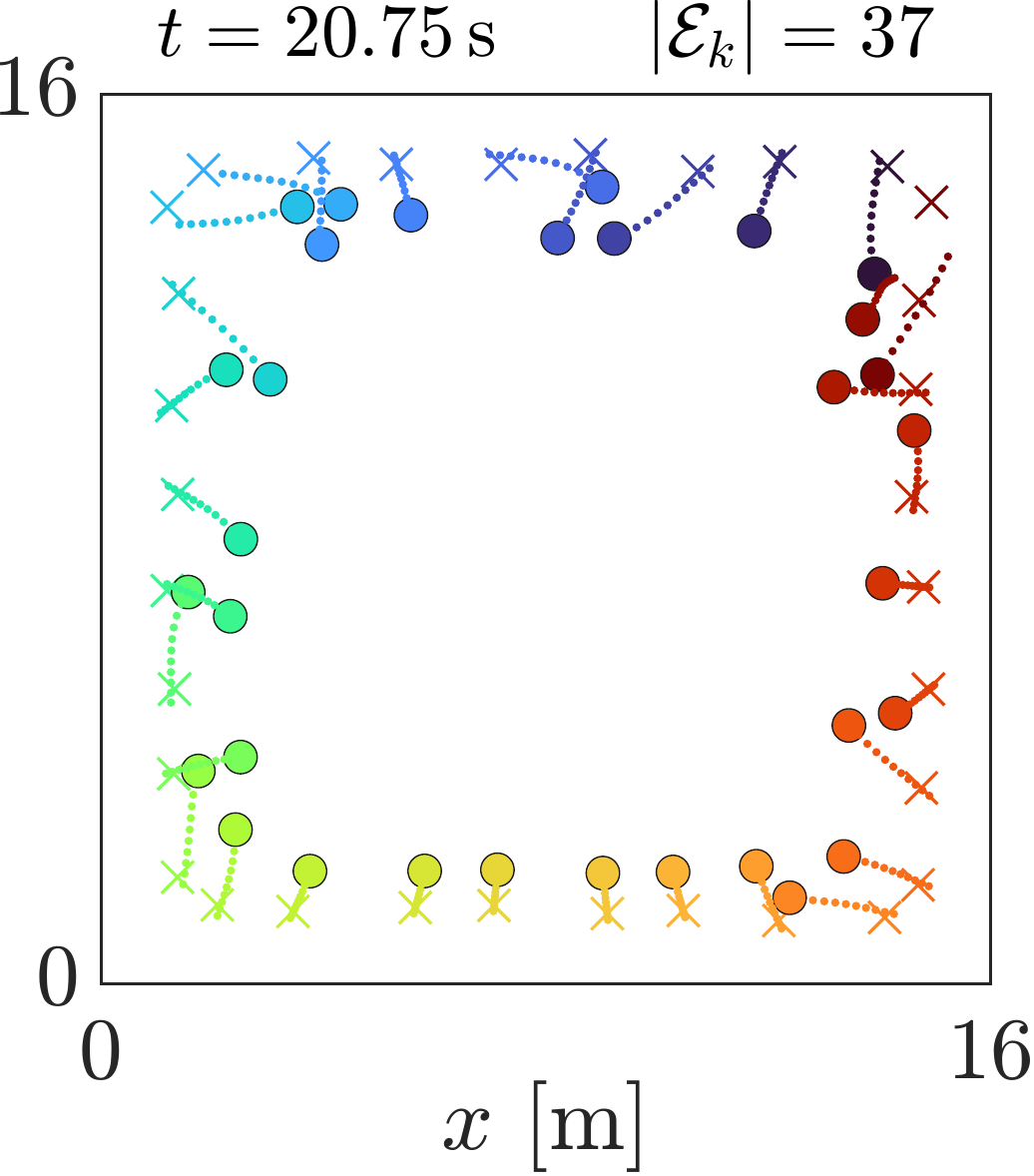}}

\caption{Representative snapshots from one distributed $32$-robot crossing.
Circles denote current robot positions, crosses the assigned goals,
and dotted sequences the predicted trajectories after ADMM termination.
Each panel reports the time and the number of active edges
$|\mathcal{E}_k|$.}
\label{fig:snapshots32}
\end{figure*}

Figure~\ref{fig:snapshots32} illustrates a representative $32$-robot crossing, showing the evolution of predicted trajectories and active interactions as the fleet approaches, negotiates, and leaves the central crossing region.

\subsection{Scalability and interaction evolution}
\label{subsec:scalability}

Table~\ref{tab:scalability_results} summarizes the closed-loop and computational performance of the centralized and distributed architectures across the three fleet sizes. The centralized formulation includes all $M(M-1)/2$ pairwise collision couplings at every MPC sample, whereas the distributed formulation includes only those selected by the predicted-conflict graph.

\begin{table*}[t]
\centering
\caption{Closed-loop and computational performance over 20 randomized realizations per fleet size.}
\label{tab:scalability_results}
\footnotesize
\setlength{\tabcolsep}{8.2pt}

\begin{tabular}{lcc@{}p{20pt}@{}cc@{}p{20pt}@{}cc}
\toprule

& \multicolumn{2}{c}{$M=8$}
&
& \multicolumn{2}{c}{$M=16$}
&
& \multicolumn{2}{c}{$M=32$} \\

\cmidrule(lr){2-3}
\cmidrule(lr){5-6}
\cmidrule(lr){8-9}

\textbf{KPI}
& \textbf{Dist.} & \textbf{Cent.}
&
& \textbf{Dist.} & \textbf{Cent.}
&
& \textbf{Dist.} & \textbf{Cent.} \\
\midrule

Mean MPC time [ms]
& $1.36$ & $12.73$
&
& $2.50$ & $126.92$
&
& $4.71$ & $1587.88$ \\

90th-percentile MPC time [ms]
& $3.00$ & $15.53$
&
& $3.96$ & $154.45$
&
& $6.71$ & $2030.52$ \\

Worst MPC time [ms]
& $6.13$ & $19.82$
&
& $8.67$ & $214.55$
&
& $14.52$ & $2306.19$ \\

Mean active edges / total pairs
& $9.72/28$ & $28/28$
&
& $27.72/120$ & $120/120$
&
& $78.56/496$ & $496/496$ \\

Min. clearance w.r.t. $d_{ij}^{\min}$ [m]
& $0.003$ & $0.014$
&
& $0.006$ & $0.002$
&
& $-0.008$ & $0.005$ \\

HOCBF call rate [\%]
& $32.52$ & 0
&
& $51.67$ & 0
&
& $63.91$ & 0 \\

Total distance [m]
& $66.90$ & $65.44$
&
& $182.55$ & $183.56$
&
& $539.34$ & $552.69$ \\

Mission completion time [s]
& $13.95$ & $12.40$
&
& $18.18$ & $16.80$
&
& $25.00$ & $25.20$ \\

\bottomrule
\end{tabular}
\end{table*}

The ratio of centralized to distributed mean MPC computation time increases from $9.36\times$ for $M=8$ to $50.77\times$ for $M=16$ and $337.1\times$ for $M=32$. At $M=32$, the worst-observed distributed update takes $14.52$~ms, compared with $2.306$~s centrally; all observed distributed updates remain below $T_s=0.25$~s. Graph sparsity also increases with fleet size: the mean fractions of active pairwise couplings are $34.7\%$, $23.1\%$, and $15.8\%$, respectively.

\begin{figure}[t]
    \centering
    \includegraphics[width=\columnwidth]
    {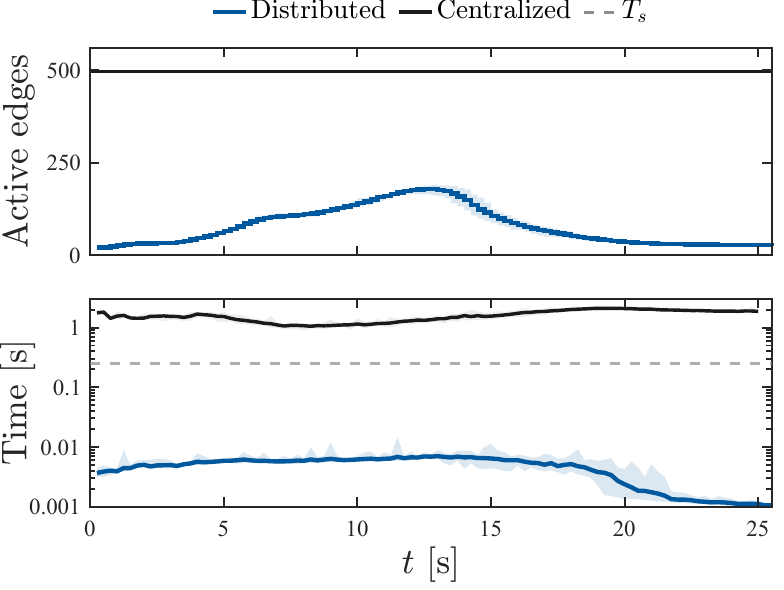}
\caption{Evolution of the $32$-robot benchmark over 20 randomized realizations. Upper panel: mean number of active edges in the distributed graph and the $496$-pair centralized reference. Lower panel: mean MPC computation time on a logarithmic scale, with $T_s=0.25$~s shown for reference. Shaded regions indicate the minimum--maximum range across realizations.}
    \label{fig:scalability32}
\end{figure}

Figure~\ref{fig:scalability32} illustrates how the interaction density and computational burden evolve in the $32$-robot benchmark. As the robots approach the central interaction region, the distributed graph becomes denser and the MPC computation time increases accordingly; both decrease again as the fleet separates. Despite this transient increase, the distributed computation time remains below the sampling period throughout the measured benchmark, whereas the centralized formulation remains substantially above $T_s$ during the maneuver.

The reported computation times exclude physical inter-robot communication and synchronization overheads. For the $32$-robot case, the 90th-percentile and worst-observed distributed computation times remain well below the sampling time. This timing margin is encouraging for real-time implementation, although validation on the target hardware and characterization of communication latency, jitter, packet loss, and synchronization effects remain for future work.

\subsection{Clearance and task-level performance}

All robots reach their assigned goals in every reported scenario under both architectures. Table~\ref{tab:scalability_results} reports executed clearance relative to $d_{ij}^{\min}=r_i+r_j+d_{\mathrm{marg}}$, including the $d_{\mathrm{marg}}=0.15$~m safety margin. The centralized controller maintains positive clearance for all three fleet sizes, with worst observed values between $0.002$ and $0.014$~m.
The distributed controller maintains positive clearance relative to the prescribed safety distance for $M=8$ and $M=16$, with worst observed values of $0.003$ and $0.006$~m, respectively.

For $M=32$, small excursions into the additional $d_{\mathrm{marg}}=0.15$~m safety margin are observed at a few isolated sampling intervals. The minimum observed clearance is
$-0.008$~m, leaving $0.142$~m of separation
between the robot boundaries.  Hence, no physical overlap occurs in any of the evaluated scenarios. 
These excursions are consistent with the sampled-data implementation of the execution filter: HOCBF constraints are enforced at discrete sampling instants, whereas the applied accelerations remain constant over the sampling interval $T_s=0.25$~s. Under continuous-time HOCBF enforcement, provided that the required invariance conditions are initially satisfied and the constraints remain feasible, such excursions would be precluded. Accordingly, reducing the sampling period would allow more frequent enforcement of the HOCBF constraints, potentially reducing intersample clearance violations.

Table~\ref{tab:distributed_diagnostics} reports ADMM termination and HOCBF intervention statistics. Here, $\Delta a_i=a_i^{\mathrm{exec}}-a_i^{\mathrm{ref}}$, and the mean intervention magnitude is evaluated over updates with $\max_i\|\Delta a_i\|_2>10^{-3}$~m/s$^2$. As fleet size grows, both the mean ADMM round count and the fraction of updates reaching $q_{\max}=15$ increase; reaching the cap does not imply failure because candidate validation and HOCBF supervision still apply. HOCBF call rates rise from $32.5\%$ ($M=8$) to $63.9\%$ ($M=32$), whereas the centralized controller requires no interventions. Mean HOCBF solve times when invoked are $0.038$, $0.129$, and $0.993$~ms for $M=8,16,32$, corresponding to $0.9\%$, $2.7\%$, and $13.5\%$ of total distributed computation time when averaged over all updates. No fallback is observed.

\begin{table}[t]
\centering
\caption{Distributed-controller diagnostics over 20 randomized realizations per fleet size.}
\label{tab:distributed_diagnostics}
\footnotesize
\setlength{\tabcolsep}{1.8pt}
\renewcommand{\arraystretch}{1.05}

\begin{tabular}{@{}p{0.43\columnwidth}ccc@{}}
\toprule
\textbf{Diagnostic}
& \textbf{$M=8$}
& \textbf{$M=16$}
& \textbf{$M=32$} \\
\midrule

Mean ADMM rounds
& $7.17$
& $10.38$
& $12.10$ \\

Residual stop [\%]
& $29.1$
& $0.0$
& $0.0$ \\

Certificate stop [\%]
& $38.4$
& $55.3$
& $38.1$ \\

$q_{\max}$ stop [\%]
& $32.4$
& $44.7$
& $61.9$ \\

\midrule

HOCBF call rate [\%]
& $32.5$
& $51.7$
& $63.9$ \\

Mean $\max_i\|\Delta a_i\|_2$ [m/s$^2$]
& $0.517$
& $0.461$
& $0.431$ \\

Maximum $\|\Delta a_i\|_2$ [m/s$^2$]
& $0.908$
& $0.910$
& $0.937$ \\

HOCBF fallbacks
& $0$
& $0$
& $0$ \\

\bottomrule
\end{tabular}
\end{table}

Despite the reduced coordination graph and finite ADMM budget, task-level performance remains comparable to the centralized baseline. Total traveled distances differ by at most approximately 2.4\%, while mission completion times remain similar across all fleet sizes, with the distributed controller slightly faster for $M=32$.

\section{Conclusions}

This paper has presented a graph-sparse distributed MPC architecture combining a time-varying predicted-conflict graph, convexified collision-avoidance constraints, edge-consensus ADMM, and selective HOCBF execution filtering.

Across $20$ randomized realizations per fleet size, the $32$-robot benchmark retains only $15.8\%$ of possible pairwise couplings and reduces mean MPC computation time by $337.1\times$ relative to the centralized complete-graph baseline. The mean 90th-percentile computation time is $6.71$~ms, below $T_s=0.25$~s and leaving substantial computational headroom for communication and synchronization overheads not represented in the present benchmark. All robots reach their goals without physical overlap, with comparable traveled distances and mission times, although HOCBF interventions become more frequent as fleet size increases.

The distributed MPC architecture is evaluated in simulation as a proof of concept, demonstrating computational scalability and effective multi-robot coordination. However, formal closed-loop safety guarantees for the sampled-data implementation remain to be established, while the supervisory HOCBF filter requires centralized computation and global state information.

Future work will investigate formal sampled-data safety guarantees and distributed execution-level safety mechanisms. Robustness to sensing, model, and communication uncertainties will also be investigated, with particular attention to asynchronous and delay-aware coordination under realistic network conditions. Experiments on distributed computing nodes and robotic platforms will quantify end-to-end communication and synchronization overheads and assess real-time closed-loop performance.

\appendices

\section{Proof of Lemma~\ref{lem:jbb-projection}}\label{app:proof_lemma1}

The feasible set is a nonempty closed half-space and the objective is coercive and strictly convex. Thus a unique minimizer exists. Introduce a multiplier $\lambda\ge0$ for $b-n_{ij,\ell}\trans(z_i-z_j)\le0$. Stationarity gives
\[
z_i=y_{i,\ell}+\lambda n_{ij,\ell},\qquad
z_j=y_{j,\ell}-\lambda n_{ij,\ell}.
\]
The constraint residual is $\delta_b-2\lambda$. Primal feasibility, dual feasibility and complementary slackness are therefore satisfied by $\lambda=[\delta_b]_+/2$. Substitution gives \eqref{eq:jbb-projection-solution}. These Karush-Kuhn-Tucker (KKT) conditions are sufficient for the convex problem.

\section{Proof of Proposition~\ref{prop:jbb-convergence}}\label{app:proof_proposition1}

Using the sequences in \eqref{eq:state_input}, introduce the auxiliary stacked vectors
\begin{equation}
\begin{aligned}
\chi_i&=\col(\xi_{i,0},\ldots,\xi_{i,N},u_{i,0},\ldots,u_{i,N-1}),\\
\chi&=\col_{i\in\mathcal I}(\chi_i),\\
z&=\col_{e=(i,j)\in\mathcal E_k}
       \bigl(\vect(Z_i^e),\vect (Z_j^e)\bigr),\\
\omega&=\col_{e=(i,j)\in\mathcal E_k}
       \bigl(\vect(W_i^e),\vect(W_j^e)\bigr).
\end{aligned}
\label{eq:jbb-stacked-variables}
\end{equation}
Here, $\vect(\cdot)$ denotes column-wise matrix vectorization.
The same fixed edge and endpoint ordering is used in $z$ and $\omega$. Define the linear extraction-and-replication operator $S$ by
\begin{equation}
S\chi=\col_{e=(i,j)\in\mathcal E_k}
       \bigl(\vect(P_i),\vect(P_j)\bigr).
\label{eq:jbb-replication}
\end{equation}
Thus $S\chi-z=0$ is precisely the collection of consensus equalities in \eqref{eq:consensus_problem}.

Let $\mathcal X$ be the set of stacked vectors $\chi$ whose associated sequences satisfy $(\xi_i,u_i)\in\mathcal F_i$ for every robot. With the indicator convention of \eqref{eq:consensus_problem}, define
\begin{equation}
\begin{aligned}
f(\chi)&=\sum_{i\in\mathcal I}J_i(\xi_i,u_i)+\ind_{\mathcal X}(\chi),\\
g(z)&=\sum_{e=(i,j)\in\mathcal E_k}\ind_{\mathcal H_e}(Z_i^e,Z_j^e).
\end{aligned}
\label{eq:jbb-split-functions}
\end{equation}
Then, \eqref{eq:consensus_problem} is equivalently
\begin{equation}
\min_{\chi,z}\ f(\chi)+g(z)
\quad\text{subject to}\quad S\chi-z=0.
\label{eq:jbb-stacked-consensus}
\end{equation}
Both functions in \eqref{eq:jbb-split-functions} are proper, closed and convex.

The operator $S$ need not have full column rank. To apply the convergence theorem without assuming otherwise, define
\begin{align}
    \mathcal X(\zeta) & =\{\chi\in\mathcal X:S\chi=\zeta\},\\
F(\zeta)& =
\begin{cases}
\displaystyle\min_{\chi\in\mathcal X(\zeta)} f(\chi),
       &\mathcal X(\zeta)\ne\varnothing,\\
+\infty,&\mathcal X(\zeta)=\varnothing.
\end{cases}
\label{eq:jbb-marginal}
\end{align}
The auxiliary vector $\zeta$ specifies replicated predicted positions. The set $\mathcal X(\zeta)$ contains all locally feasible state and input sequences yielding those positions, and $F(\zeta)$ is their minimum total cost. Every nonempty $\mathcal X(\zeta)$ is compact, so the minimum is attained. Convexity of $F$ follows by taking convex combinations of feasible minimizers. 

To verify lower semicontinuity, let $\zeta^m \to \zeta$ and set $L=\liminf_{m\to\infty}F(\zeta^m)$. If $L=+\infty$, the inequality $F(\zeta)\le L$ is immediate. Otherwise, since the quadratic cost is bounded below on the compact set $\mathcal X$, $L$ is finite. Choose a subsequence
$\{\zeta^{m_j}\}$ such that $F(\zeta^{m_j})\to L$ and $F(\zeta^{m_j})<+\infty$.
For each $j$, select a minimizer $\chi^{m_j}\in\mathcal X(\zeta^{m_j})$, so that
$f(\chi^{m_j})=F(\zeta^{m_j})$.

Compactness of $\mathcal X$ guarantees a further subsequence, denoted by the same indices, converging to some $\bar\chi\in\mathcal X$. Since $S$ is linear and continuous, $S\bar\chi=\zeta$, hence $\bar\chi\in\mathcal X(\zeta)$. Continuity of the quadratic cost on $\mathcal X$ therefore
gives
$$
F(\zeta)\le f(\bar\chi)
=\lim_{j\to\infty}f(\chi^{m_j})
=\lim_{j\to\infty}F(\zeta^{m_j})
=L.
$$
Thus $F$ is lower semicontinuous and, together with the properties established above, proper, closed and convex.

Consequently, \eqref{eq:jbb-stacked-consensus} is equivalent to
\begin{equation}
\min_{\zeta,z}\ F(\zeta)+g(z)
\quad\text{subject to}\quad\zeta-z=0.
\label{eq:jbb-reduced-consensus}
\end{equation}
The saddle point assumed for \eqref{eq:consensus_problem} induces a saddle point of \eqref{eq:jbb-reduced-consensus}, with the same multiplier for the consensus equality.

Stacking the updates in \eqref{eq:node_update} gives
\begin{equation}
\chi^{q+1}\in\arg\min_\chi
\left\{f(\chi)+\frac\rho2\|S\chi-z^q+\omega^q\|_2^2\right\}.
\label{eq:jbb-stacked-local}
\end{equation}
For fixed $S\chi$, the quadratic term is constant. Hence \eqref{eq:jbb-stacked-local} selects a minimizer over $\mathcal X(S\chi^{q+1})$, and $\zeta^{q+1}=S\chi^{q+1}$ solves the $\zeta$-update of \eqref{eq:jbb-reduced-consensus}. The matrix updates in \eqref{eq:admm_relaxation} and \eqref{eq:dual_update}, together with Lemma~\ref{lem:jbb-projection}, become
\begin{equation}
\begin{aligned}
\widehat\zeta^{q+1}&=\alpha\zeta^{q+1}+(1-\alpha)z^q,\\
z^{q+1}&=\arg\min_z\left\{g(z)+\frac\rho2
          \|\widehat\zeta^{q+1}-z+\omega^q\|_2^2\right\},\\
\omega^{q+1}&=\omega^q+\widehat\zeta^{q+1}-z^{q+1}.
\end{aligned}
\label{eq:jbb-reduced-updates}
\end{equation}
Equations \eqref{eq:jbb-stacked-local}--\eqref{eq:jbb-reduced-updates} are exactly relaxed two-block ADMM for \eqref{eq:jbb-reduced-consensus}. The classical convergence result for fixed $\rho>0$ and $\alpha\in(0,2)$ yields $\zeta^q-z^q\to0$ and $F(\zeta^q)+g(z^q)\to J_k^\star$; see \cite[Sections 3.2 and 3.4.3]{boyd2011}.

By \eqref{eq:jbb-replication}, the norm $\|\zeta^q-z^q\|_2$ is exactly the primal residual $r_p^q$ defined in \eqref{eq:admm-primal-residual}. Moreover, \eqref{eq:jbb-stacked-local} implies $f(\chi^q)=F(\zeta^q)$, and the projection in \eqref{eq:jbb-reduced-updates} gives $g(z^q)=0$, for $q\ge1$. This proves $r_p^q\to0$ and \eqref{eq:jbb-objective-convergence}. Any accumulation point $\chi^\star$ lies in the compact set $\mathcal X$. Along a subsequence converging to $\chi^\star$, vanishing consensus error implies $z^q\to S\chi^\star$. Closedness of the sets in \eqref{eq:edge_set} makes that limit feasible for \eqref{eq:consensus_problem}, and continuity of the total cost makes it optimal.

Finally, \eqref{eq:prox_weight} gives $w_\ell=w_0\beta^\ell>0$ for every input stage. The proximal term in \eqref{eq:local_cost} is strictly convex in the stacked accelerations and all other terms are convex. Two optimizers of \eqref{eq:consensus_problem} with different acceleration sequences would therefore have a feasible midpoint with strictly smaller cost, a contradiction. All optimizers share the same acceleration sequence. Compactness and the optimality of every accumulation point then imply convergence of the iterated acceleration sequences to that value. Applying the triangle inequality to successive first controls proves \eqref{eq:jbb-control-increment}.

\section{Proof of Lemma~\ref{lem:jbb-separation}}\label{app:proof_lemma2}

For a fixed edge $e=(i,j)$ and stage $\ell$, introduce the owner--copy errors
\begin{equation}
\varepsilon_{i,\ell}^{e,q}=p_{i,\ell}^q-z_{i,\ell}^{e,q},
\qquad
\varepsilon_{j,\ell}^{e,q}=p_{j,\ell}^q-z_{j,\ell}^{e,q}.
\label{eq:jbb-owner-copy-errors}
\end{equation}
Using \eqref{eq:jbb-copy-margin} and \eqref{eq:jbb-owner-copy-errors},
\begin{equation}
\begin{aligned}
n_{ij,\ell}\trans(p_{i,\ell}^q-p_{j,\ell}^q)-d_{ij}^{\min}
&=s_{e,\ell}^q+n_{ij,\ell}\trans\varepsilon_{i,\ell}^{e,q}
             -n_{ij,\ell}\trans\varepsilon_{j,\ell}^{e,q}\\
&\ge s_{e,\ell}^q-\|\varepsilon_{i,\ell}^{e,q}\|_2
                  -\|\varepsilon_{j,\ell}^{e,q}\|_2\\
&=\eta_{e,\ell}^q.
\end{aligned}
\label{eq:jbb-certificate-expansion}
\end{equation}
Inequality follows from Cauchy--Schwarz and the unit norm of $n_{ij,\ell}$; the final equality is exactly \eqref{eq:finite_round_certificate}. Since
$n_{ij,\ell}\trans(p_{i,\ell}^q-p_{j,\ell}^q)\le\|p_{i,\ell}^q-p_{j,\ell}^q\|_2$,
this proves the first two inequalities in \eqref{eq:jbb-edge-bound}.

The two column errors in \eqref{eq:jbb-owner-copy-errors} are among the terms whose squared norms are summed in \eqref{eq:admm-primal-residual}. Therefore,
\begin{align}\label{eq:jbb-error-residual-bound}
\|\varepsilon_{i,\ell}^{e,q}\|_2+\|\varepsilon_{j,\ell}^{e,q}\|_2
&\le\sqrt{2}\,
 \sqrt{\|\varepsilon_{i,\ell}^{e,q}\|_2^2+
       \|\varepsilon_{j,\ell}^{e,q}\|_2^2} \notag \\
&\le\sqrt{2}\,
 \sqrt{\|P_i^q-Z_i^{e,q}\|_F^2+
       \|P_j^q-Z_j^{e,q}\|_F^2}\notag \\
&\le\sqrt{2}\,r_p^q.
\end{align}

Substituting \eqref{eq:jbb-error-residual-bound} into the expression for $\eta_{e,\ell}^q$ in \eqref{eq:jbb-certificate-expansion} proves the last inequality in \eqref{eq:jbb-edge-bound}. Taking minima over the same active edges and stages as in \eqref{eq:worst_case_certif} proves \eqref{eq:jbb-global-bound}. Finally, if $\eta_{\min}^q\ge\eta_{\mathrm{cert}}$, every $\eta_{e,\ell}^q$ is at least that threshold; \eqref{eq:jbb-edge-bound} then yields \eqref{eq:jbb-certified-owner-separation}.

\balance

\begin{IEEEbiography}[{\includegraphics[width=1in,height=1.25in,clip,keepaspectratio]{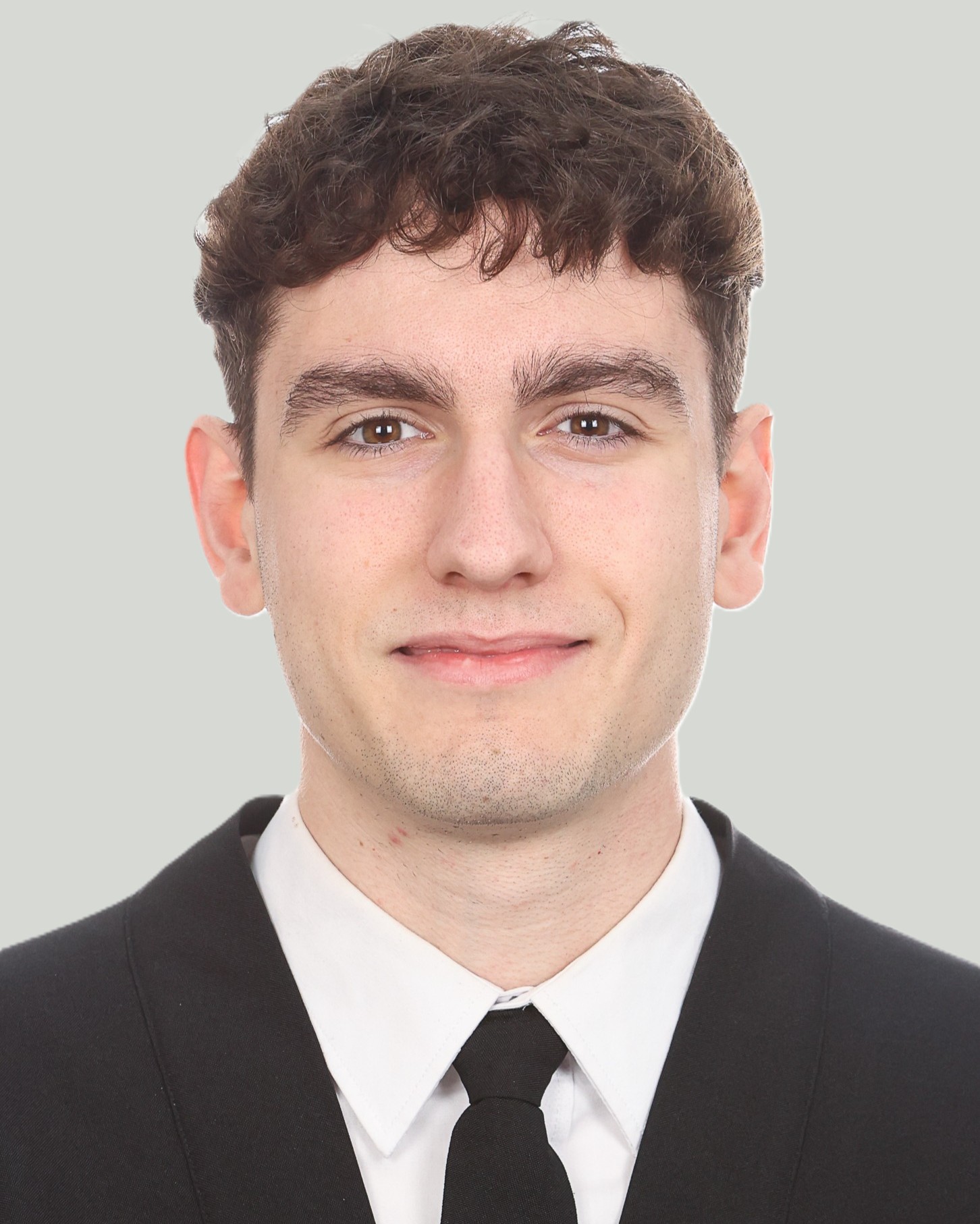}}]{Josep Borràs-Borràs}
received the B.Sc. degree in Industrial Technologies and Economic Analysis from Universitat Politècnica de Catalunya (UPC) and Universitat Pompeu Fabra (UPF), Barcelona, Spain, in 2025. He is currently pursuing the M.Sc. degree in Industrial Engineering at UPC, specializing in Automation and Robotics. His research interests include model predictive control, distributed optimization, multi-robot systems, 
and motion planning.
\end{IEEEbiography}

\begin{IEEEbiography}[{\includegraphics[width=1in,height=1.25in,clip,keepaspectratio]{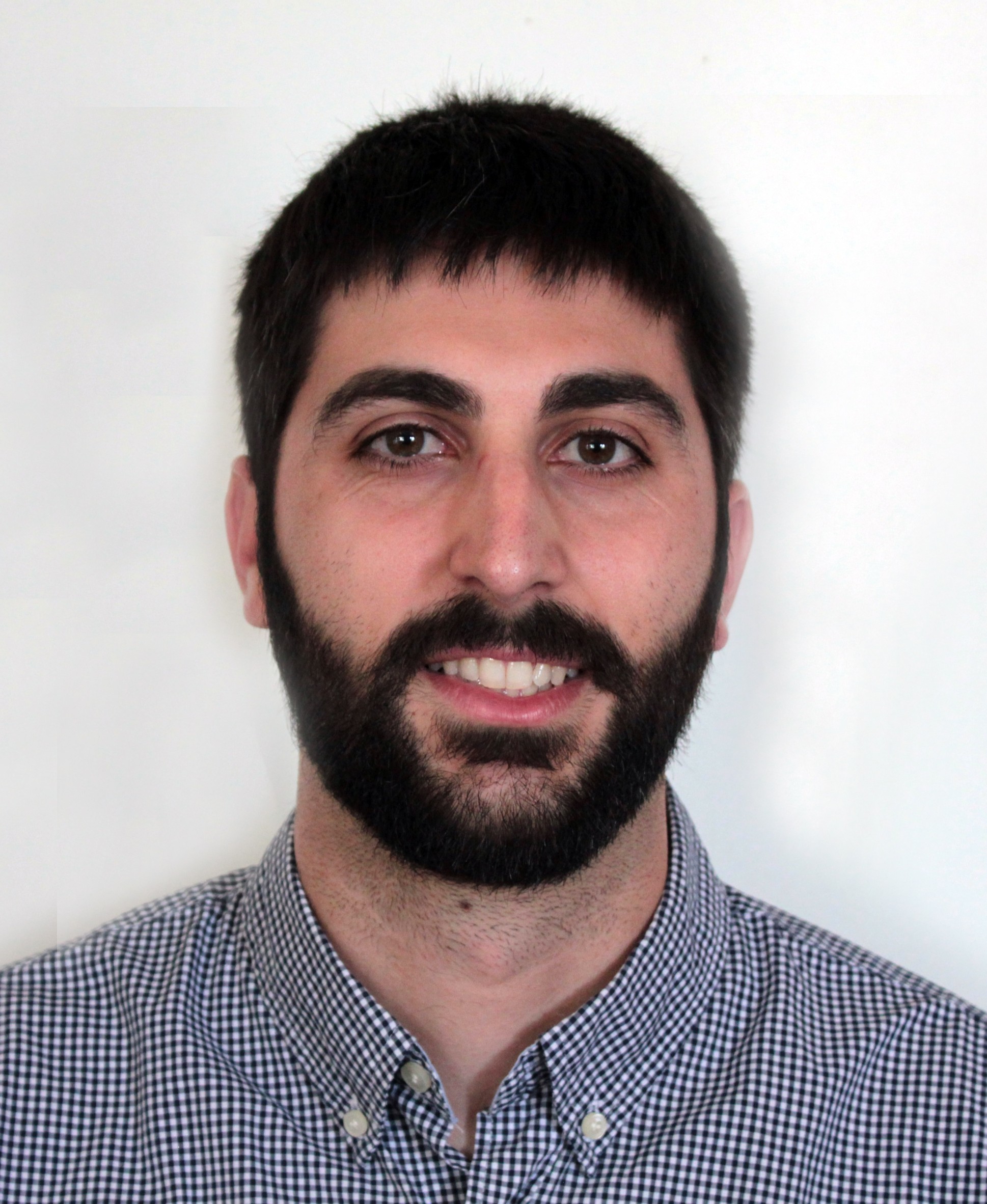}}]{Pablo Segovia} received the B.Sc./M.Sc. degree in Industrial Engineering in 2015 and the joint Ph.D. degree in Automatic Control in 2019 from Universitat Politècnica de Catalunya (UPC), Barcelona, Spain, and IMT Nord Europe, France. He held postdoctoral research appointments at IMT Nord Europe (2019–2020), TU Delft (2020–2023), and UPC (2023), where he is currently a Beatriz Galindo Fellow with the Automatic Control Department. His research interests include the control of large-scale network systems, water resources systems and intelligent transportation systems.
\end{IEEEbiography}

\begin{IEEEbiography}[{\includegraphics[width=1in,height=1.25in,clip,keepaspectratio]{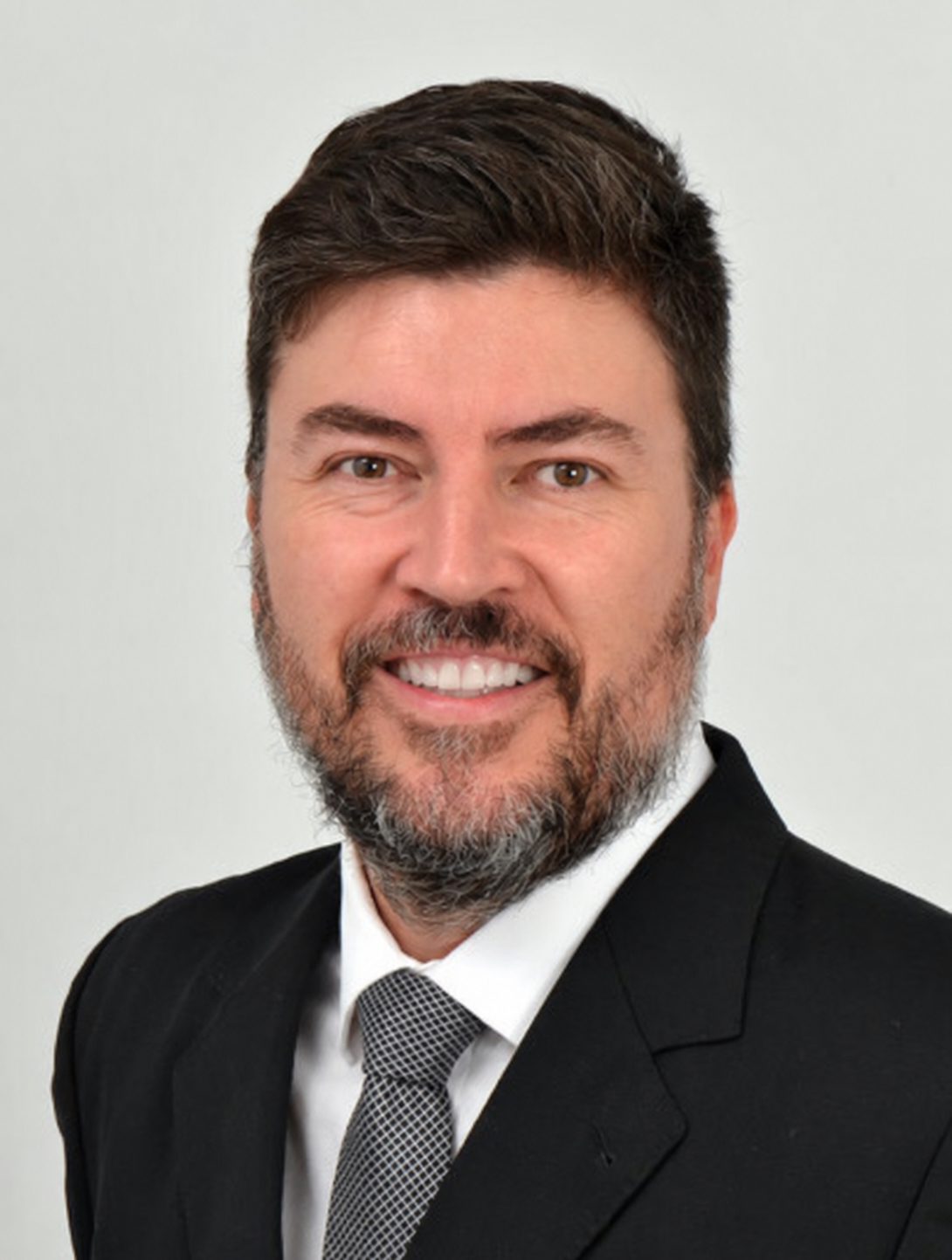}}]{Carlos Ocampo-Martinez} (SM'13) received his Ph.D. in Control Engineering from the Universitat Politècnica de Catalunya (UPC), Barcelona, Spain, in 2007. He is currently a Full Professor at UPC, specializing in Automatic Control and Model Predictive Control. He has authored or co-authored over 300 scientific publications and seven Springer books. His research focuses on model predictive control, large-scale systems, non-centralized control, game-theoretic approaches, and industrial applications in water, energy, and smart manufacturing. He currently serves as Associate Editor for four leading international journals, including IEEE Transactions on Control of Network Systems, IEEE Open Journal of Control Systems, IEEE Control Systems Letters, and Journal of Process Control.
\end{IEEEbiography}

\end{document}